\documentclass[compsoc]{IEEEtran}
\usepackage{lineno,hyperref}
\usepackage{times}
\usepackage{url}
\usepackage{latexsym} 
\usepackage{xcolor}
\usepackage{microtype}
\usepackage{graphicx}
\usepackage{booktabs}
\usepackage{microtype}
\usepackage{latexsym}
\usepackage{stfloats}
\usepackage{multirow}
\usepackage{enumitem}
\usepackage{amsmath}
\usepackage{algorithm}
\usepackage{booktabs,fixltx2e}
\usepackage{amssymb,mathtools}
\usepackage{eqnarray}
\usepackage{hyperref}
\usepackage{microtype}
\usepackage{subfig}
\usepackage{lscape}
\usepackage{arydshln}
\usepackage{orcidlink}
\usepackage{setspace}

\modulolinenumbers[5]
\ifCLASSINFOpdf 
\else 
\fi 

\begin{document}

%% =*=*=*=*=*=*=*=*=*=*=*=*=*=*=*=*=*=*=*=*=*=*=*=*=*=*=*=*=*=*
\title{VISTANet: VIsual Spoken Textual Additive Net for Interpretable Multimodal Emotion Recognition}
%Interpretable Multimodal Emotion Recognition: Addressing Data Limitations with a New Dataset
%Hybrid Fusion based Interpretable Multimodal Emotion Recognition with Limited Labelled Data

\author{
    Puneet~Kumar\textsuperscript{\dag}
        \orcidlink{0000-0002-4318-1353},
        \IEEEmembership{Member,~IEEE},
    Sarthak~Malik\textsuperscript{\dag}
        \orcidlink{0000-0001-5224-1445},
    Balasubramanian~Raman\textsuperscript{*}
        \orcidlink{0000-0001-6277-6267},
        \IEEEmembership{Senior~Member,~IEEE}, and 
    Xiaobai Li\textsuperscript{*}
        \orcidlink{0000-0003-4519-7823},
        \IEEEmembership{Senior~Member,~IEEE}
        
    \thanks{\textsuperscript{\dag} Contributed Equally.}
    \thanks{\textsuperscript{*} Corresponding Author.}
    \thanks{P. Kumar is with the Center for Machine Vision and Signal Analysis, University of Oulu, Finland. Email: puneet.kumar@oulu.fi}
    \thanks{S. Malik and B. Raman are with Indian Institute of Technology Roorkee, India. E-mail: sarthak\_m@mt.iitr.ac.in and bala@cs.iitr.ac.in}
    \thanks{X. Li is with the State Key Laboratory of Blockchain and Data Security, Zhejiang University, Hangzhou, China and the Center for Machine Vision and Signal Analysis, University of Oulu, Finland. E-mail: xiaobai.li@zju.edu.cn}

    % \thanks{Manuscript received Nov 6, 2023 ...}
}

% The paper headers
%\markboth{IEEE Transactions on Affective Computing, Vol.~01, No.~1, May~2025}
\markboth{Journal of \LaTeX\ Class Files,~Vol.~14, No.~8, August~2015}%
{Shell \MakeLowercase{\textit{et al.}}: Bare Demo of IEEEtran.cls for IEEE Journals}

% =*=*=*=*=*=*=*=*=*=*=*=*=*=*=*=*=*=*=*=*=*=*=*=*=*=*=*=*=*=*
\IEEEtitleabstractindextext{
    \begin{abstract} 
    This paper proposes a multimodal emotion recognition system, VIsual Spoken Textual Additive Net (VISTANet), to classify emotions reflected by input containing image, speech, and text into discrete classes. A new interpretability technique, K-Average Additive exPlanation (KAAP), has been developed that identifies important visual, spoken, and textual features leading to predicting a particular emotion class. The VISTANet fuses information from image, speech, and text modalities using a hybrid of intermediate and late fusion. It automatically adjusts the weights of their intermediate outputs while computing the weighted average. The KAAP technique computes the contribution of each modality and corresponding features toward predicting a particular emotion class. To mitigate the insufficiency of multimodal emotion datasets labelled with discrete emotion classes, we have constructed the IIT-R MMEmoRec dataset consisting of images, corresponding speech and text, and emotion labels (`angry,' `happy,' `hate,' and `sad'). The VISTANet has resulted in an overall emotion recognition accuracy of 80.11\% on the IIT-R MMEmoRec dataset using visual, spoken, and textual modalities, outperforming single or dual-modality configurations. The code and data can be accessed at \href{https://github.com/MIntelligence-Group/MMEmoRec}{\underline{github.com/MIntelligence-Group/MMEmoRec}}.
    \end{abstract}
    
    \begin{IEEEkeywords}
    Affective Computing, Emotion and Sentiment Analysis, Speech-Text-Image Signals, Information Fusion, Interpretable AI.
    \end{IEEEkeywords}
}

% make the title area
\maketitle
\IEEEpeerreviewmaketitle

%====================================
%-*-*-*-*-*- Introduction -*-*-*-*-*-
%====================================
\section{Introduction}\label{sec:intro}
\IEEEPARstart{T}{he} multimedia data has grown in the last few years, leading multimodal emotion analysis to emerge as an important research trend~\cite{baltruvsaitis2018multimodal}. It is used in various applications such as cognitive psychology, automated identification, intelligent devices, and human-machine interface~\cite{ezzameli2023emotion}. Humans portray emotions through various modalities such as images, speech, and text~\cite{muszynski2019recognizing}. Utilizing the multimodal information from them could increase the performance of emotion recognition~\cite{wang2023mgeed}. Researchers have performed emotion recognition by analyzing visual, spoken, and textual information separately ~\cite{shingjergji2022interpretable, palash2023emersk, majumder2019dialoguernn}. Multimodal emotion recognition using two modalities has been explored; however, it is yet to be fully explored using all three \cite{wang2023mgeed}. Moreover, most existing multimodal approaches do not focus on interpreting the internal workings of their emotion recognition systems \cite{malik2021towards}. 

Multimodal emotion recognition faces the unavailability of sufficient labelled data for training. Moreover, real-life multimodal data contains generic images with facial, human, and non-human objects, yet most existing datasets include only facial images~\cite{busso2008iemocap}. In this context, perceived emotions, recognized by observers in multimodal content, differ from induced emotions which are reflected by the subjects themselves~\cite{tian2017recognizing}. This paper focuses on perceived emotions because they more accurately reflect how individuals interact with and interpret real-world stimuli. To address this, we propose the IIT-R MMEmoRec dataset, which captures a key aspect of real-world interactions by including various image types beyond facial expressions. Unlike a few multimodal datasets that contain generic images but only offer positive, negative, and neutral sentiment labels~\cite{gaspar2019multimodal, Vadicamo_2017_ICCVW}, the IIT-R MMEmoRec dataset includes generic images, corresponding speech utterances, text transcripts, and discrete labels: `happy,’ `sad,’ `hate,’ and `anger.’

This paper proposes an interpretable multimodal emotion recognition system, VIsual Spoken Textual Additive Net (VISTANet), which combines features from images, speech, and text using a hybrid of intermediate and late fusion techniques. It utilizes a combination of complex pre-trained models along with simpler models. This configuration helps the simpler model learn and adapt by leveraging the robust knowledge of the pre-trained model, thus improving overall integration and responsiveness. A novel interpretability technique, K-Average Additive exPlanation (KAAP), has also been developed to identify important visual, spoken, and textual features predicting specific emotion classes. It is used to automatically adjust intermediate outputs and compute the weighted average without human input.%The modality weights for the fusion are computed using the grid search.

{The IIT-R MMEmoRec dataset includes Set A, containing highly confident samples, and an additional Set B, with samples subjected to less strict filtering. The VISTANet has achieved emotion recognition accuracies of 95.99\%, 75.13\%, and 80.11\% for Set A, Set B, and the overall dataset, respectively. In the ablation studies on Set A,} an accuracy of 81.95\% was observed for emotion recognition combining speech and text. The combinations of image and text, and speech and image, achieved accuracies of 86.40\% and 84.66\%, respectively. VISTANet outperformed these configurations by achieving a 95.99\% accuracy when integrating all three modalities. Furthermore, the KAAP technique identifies the contributions of each modality and its features. The major contributions of this work are as follows.\vspace{-.018in}

\begin{itemize}
\item A hybrid-fusion-based novel interpretable multimodal emotion recognition system, VISTANet, has been proposed to classify an input containing an image, corresponding speech, and text into discrete emotion classes. \vspace{.02in}
\item A novel interpretability technique, KAAP, has been developed to identify each modality's significance and the key image, speech, and text features contributing to recognizing emotions. 
\item A large-scale dataset, `IIT-R MMEmoRec dataset' containing images, speech utterances, text transcripts, and emotion labels has been constructed.%\footnote{\noindent The IIT-R MMEmoRec dataset can be accessed at \href{https://github.com/MIntelligence-Group/MMEmoRec}{\underline{github.com/MIntelligence-Group/MMEmoRec}}.}\vspace{-.2in}
\end{itemize} 

%Further in this paper, Section~\ref{sec:lr} reviews related works. The proposed dataset, system, and interpretability technique, as well as the dataset construction procedure, are detailed in Section~\ref{sec:method}. Sections~\ref{sec:experiments} and \ref{sec:results} cover the experiments and their outcomes, and the paper concludes with Section~\ref{sec:conclusion}.

%=======================================
%-*-*-*-*-*- Related Works -*-*-*-*-*-
%=======================================
\section{Related works}\label{sec:lr} 
%\noindent This Section surveys the existing literature on speech \& image emotion recognition and deep neural network interpretability. 
\subsection{Unimodal Emotion Recognition}
In unimodal emotion recognition, individual modalities like speech, text, and images are utilized to detect emotions. Speech Emotion Recognition (SER) systems traditionally extract speech features such as cepstrum coefficients, voice tone, prosody, and pitch, key for identifying emotions~\cite{jing2023deep}. These features help categorize high-key emotions like happiness and anger from low-key ones such as sadness and despair~\cite{lorenzo2015emotion}. However, the manual crafting of acoustic features and difficulties in parameter estimation pose challenges in developing robust SER systems~\cite{uden2022mrec}. Recent advances using spectrogram features and attention mechanisms have enhanced SER's effectiveness~\cite{dai2019learning,kimfocus2023}, with CNNs for spectrogram processing~\cite{mao2014learning} and RNN-based techniques showing promising results~\cite{majumder2019dialoguernn}.

Text Emotion Recognition (TER) analyzes emotions portrayed by transcripts from online platforms like YouTube, Facebook, and Twitter~\cite{ma2023transformer}. Attention mechanisms using graphs~\cite{li2023ga2mif}, transformer models~\cite{ma2023transformer}, word embedding techniques from tweets~\cite{abubakar2022explainable}, and graph network-based multimodal fusion~\cite{li2023graphcfc} are leading methods for TER. Sequence-based CNNs and the integration of semantic and emotional information enhance the modelling of textual emotions~\cite{shrivastava2019effective,batbaatar2019semantic}. Image Emotion Recognition (IER) primarily uses facial expressions, benefiting from techniques like face localization, micro-expression analysis, and landmark tracking, and both traditional and advanced deep learning approaches~\cite{shingjergji2022interpretable,corneanu2016survey}. Despite advancements, IER faces challenges from deep learning techniques, underscoring the need for continued research~\cite{shingjergji2022interpretable}.
	
%\noindent \textit{Traditional IER approaches}: 
%Feature-Based Semantic Image Analysis -- IER was initially done using low-level features such as shape, edge, and color~\cite{hanjalic2006extracting}. Joshi et al.~\cite{joshi2011aesthetics} found that mid-level features such as optical balance and composition contribute to image aesthetics. Zhao et al.~\cite{zhao2014exploring} applied mid-level features for image emotion classification. In another work, Machajdik et al.~\cite{machajdik2010affective} used the semantic content of the images for emotion analysis. However, the methods mentioned above use handcrafted features which are not likely to accommodate all low-level features, mid-level features, and image semantics.

%\noindent \textit{Deep Learning based IER methods}: 
%Object classification, image recognition, and other computer vision tasks have successfully used CNN~\cite{krizhevsky2012imagenet, long2015fully}. It extracts the visual features in an end-to-end manner without human intervention. However, the currently used CNN methods face difficulty extracting mid and low-level image features, which are required to identify emotion-related information in the images~\cite{rao2019learning} precisely. They also need large-scale well-labelled image datasets for training. Such datasets are not abundant, and emotion labelling is subjected to human variations.
	
\subsection{Multimodal Emotion Recognition}
Emotion analysis using a single modality may not fully capture the emotional context \cite{wang2023mgeed}. Various modalities have distinct statistical properties, and understanding the inter-relationships between them is crucial for recognizing complex emotions \cite{ezzameli2023emotion}, leading researchers to focus on multimodal emotion analysis \cite{bhattacharya2021exploring}. Existing approaches include interpreting emotions from speech and text using activation vectors \cite{asokan2022interpretability}, and acoustic and textual analyses \cite{makiuchi2021multimodal}. Dual RNNs extract speech and text information for emotion recognition \cite{yoon2018multimodal}, with transformers-based models being fine-tuned to enhance performance \cite{siriwardhana2020jointly}. Systems employing multiple modalities often utilize information fusion and end-to-end approaches for effective integration \cite{poria2016fusing, tzirakis2017end}. Studies involving visual and textual data have explored semantic reasoning networks \cite{zhu2022multimodal} and multi-task architectures designed to address missing modalities \cite{xie2023multimodal}. Additionally, co-memory-based networks have been developed for sentiment recognition \cite{xu2018co}. The adoption of pre-trained transformers has significantly advanced multimodal emotion recognition (MER) efforts \cite{shayaninasab2024multi, mai2022multimodal}.

Multimodal emotion recognition emphasizes integrating feature extraction and fusion techniques across visual, speech, and textual modalities. Transformer-based models play a central role due to their ability to handle complex intermodal interactions. Ma et al. introduced a self-distilling transformer model enhancing emotion recognition accuracy in conversational contexts \cite{ma2023transformer}, demonstrating transformers' significant performance enhancement by leveraging multimodal data. Zhang et al. provided a systematic review of deep learning approaches for MER, underscoring the transformative impact of integrating modalities through advanced architectures like transformers \cite{zhang2023deep}. Fan et al. developed transformer-based networks for depression detection, highlighting the adaptability of this technology in affective computing \cite{fan2024transformer}. Advancements continue with methods integrating fused speech and visual features \cite{wei2022fv2es, hu2022unimse}, modality-specific frameworks \cite{patamia2023multimodal}, knowledge-embedded models for deeper analysis \cite{zhao2022memobert, zheng2023two}, and emotion analysis using self-supervised learning and feature correlation analysis \cite{aytar2016soundnet, guanghui2021multi}.

\subsection{Explainable and Interpretable Emotion Analysis}
Explainability describes an algorithm’s mechanism for a specific output, while interpretability concerns understanding a model’s output in context and its functional design \cite{kumartowards, kumar2025multimodal}. The opaque nature of deep learning has driven the emergence of explainable AI \cite{lundberg2017unified}. Ribeiro et al. \cite{fazi2020beyond} developed perturbation-based methods to identify which input components drive outputs, and Shrikumar et al. \cite{shrikumar2017learning} traced individual neuron contributions via backpropagation. Existing interpretability techniques fall into three categories—attribution-based, perturbation-based, and backpropagation-based methods with the latter two being subsets of attribution approaches. These attribution-based methods focus on local interpretability of individual instances. Attribution methods compute feature relevance via Shapley values \cite{SHAPley1953value}, which underpin the SHAP framework \cite{lundberg2017unified,malik2021towards}. However, exact Shapley computation is combinatorial, requiring evaluation of \(2^n\) subsets for \(n\) features, motivating approximations such as KernelSHAP \cite{lundberg2017unified} and sampling-based Shapley values \cite{castro2009polynomial}. {Another interpretability approach, DeepSHAP builds on DeepLIFT and requires direct access to internal neuron activations and gradients—making it model-dependent and necessitating separate explainers per modality \cite{shrikumar2017learning}. Such gradient-based attributions cannot be uniformly applied to transformer-based text backbones like BERT \cite{bolukbasi2021interpretability}.}

Perturbation techniques involve making small alterations to inputs and observing their impact on the model's behaviour \cite{fong2019understanding}. Local Interpretable Model-agnostic Explanations (LIME) \cite{ribeiro2016should} is a widely used perturbation technique that generates new data by perturbing the original instance and weights it based on proximity. Although applicable to any machine learning model, LIME's reliance on generating new data can lead to computational overhead. Backpropagation-based techniques compute attributions by iteratively backpropagating through the network. Saliency Maps \cite{simonyan2014deep} and Gradient-weighted Class Activation Map (Grad-CAM) \cite{selvaraju2017grad} are prominent examples. Grad-CAM generates highlights of important input features by focusing on the last convolutional layer, thereby offering insights into model decisions \cite{kindermans2019reliability}.

As highlighted, techniques like LIME, SHAP, and Grad-CAM have challenges: LIME is computationally costly, Grad-CAM struggles with minor input changes, and SHAP, though robust, is mainly for visual modalities. These issues led to the development of our KAAP interpretability technique for multimodal emotion recognition. Unlike traditional attention-based methods that highlight features without quantifying impact \cite{palash2023emersk, mai2022multimodal, lian2023mer}, KAAP offers a quantitative analysis of feature influence. It uses a perturbation-based method to evaluate contributions, providing precise insights into how modalities and features affect outcomes. This approach addresses limitations in attention-based models and enhances interpretability in multimodal emotion analysis.

\section{Proposed Work}\label{sec:method}
% \subsection{Problem Formulation}\label{sec:problem}
% {\color{blue} 
%     To write mathematically, three types of inputs, outputs one of 4 classes...
% }
%     Given a classification problem containing multiple modalities. The proposed architecture combines each of the modalities in a hybrid manner to give the final prediction, furthermore, given a Deep Neural Network that maps the multimodular feature space $\mathbb{X}$ to $m$ discrete labels. The proposed interpretability approach aims to calculate the importance of each modality towards predictions and the importance of each feature within each modality.  %in this case importance of each pixel. 
% }
\subsection{Data Compilation}\label{sec:proposed}
The IIT-R MMEmoRec dataset contains generic images (facial, human, non-human objects), speech utterances, text transcripts, emotion labels (`angry,' `happy,' `hate,' and `sad'), and the probability of each emotion class. In contrast, other multimodal datasets like the Interactive Emotional Dyadic Motion Capture (IEMOCAP) \cite{busso2008iemocap} and the Multimodal EmotionLines Dataset (MELD) \cite{poria2018meld} contain only facial images/videos. {Moreover, the IIT-R MMEmoRec dataset demonstrates greater diversity than IEMOCAP and MELD. It features faces in 70.64\% of its images, compared to 97.86\% in the other datasets, showing a higher variety of non-facial images. Additionally, the IIT-R MMEmoRec dataset includes 80 YOLOv8 objects \cite{YOLOv8}, with 30 different ones appearing in over 1\% of the images, whereas the IEMOCAP and MELD datasets contain 72 objects, with 18 different ones appearing in more than 1\% of the images. Furthermore, 14.48\% of its images feature unidentified YOLOv8 objects, while this figure is only 1.2\% for the other datasets, underscoring its broader diversity.} It has been constructed on top of the `Balanced Twitter for Sentiment Analysis' (B-T4SA) dataset \cite{Vadicamo_2017_ICCVW}, which contains images, text, and sentiment labels (`positive,' `negative,' `neutral'). The IIT-R MMEmoRec dataset has discrete emotion labels for image, text, and speech modalities and has been constructed as described.

% The IIT-R MMEmoRec dataset contains generic (facial, human, non-human objects) images, speech utterances, text transcripts, emotion labels (`angry,' `happy,' `hate,' and `sad'), and the probability of each emotion class. In contrast, the other multimodal datasets like Interactive Emotional Dyadic Motion Capture (IEMOCAP) \cite{busso2008iemocap} and Multimodal EmotionLines Dataset (MELD) \cite{poria2018meld}) contain only facial images/videos. {Moreover, the IIT-R MMEmoRec dataset demonstrates greater diversity than IEMOCAP and MELD. It features faces in 70.64\% of its images, as opposed to 97.86\% in the other datasets, showing a higher variety of non-facial images. Moreover, IIT-R MMEmoRec dataset includes 80 objects with 30 different ones appearing in over 1\% of the images. The IEMOCAP and MELD datasets contain 72 objects with 18 different ones appearing in more than 1\% of the images. Additionally, 14.48\% of its images feature unidentified YOLOv8 objects, while this number is 1.2\% for the other datasets, underscoring its broader diversity.} It has been constructed on top of the `Balanced Twitter for Sentiment Analysis' (B-T4SA) dataset~\cite{Vadicamo_2017_ICCVW} containing images, text, and sentiment (`positive,' `negative,' neutral) labels. The IIT-R MMEmoRec dataset has discrete emotion labels for image, text, and speech modalities and it has been constructed as follows. %MOSEI \cite{zadeh2018multi}

\begin{figure}[!t]
\centering
\includegraphics[width=.29\textwidth]{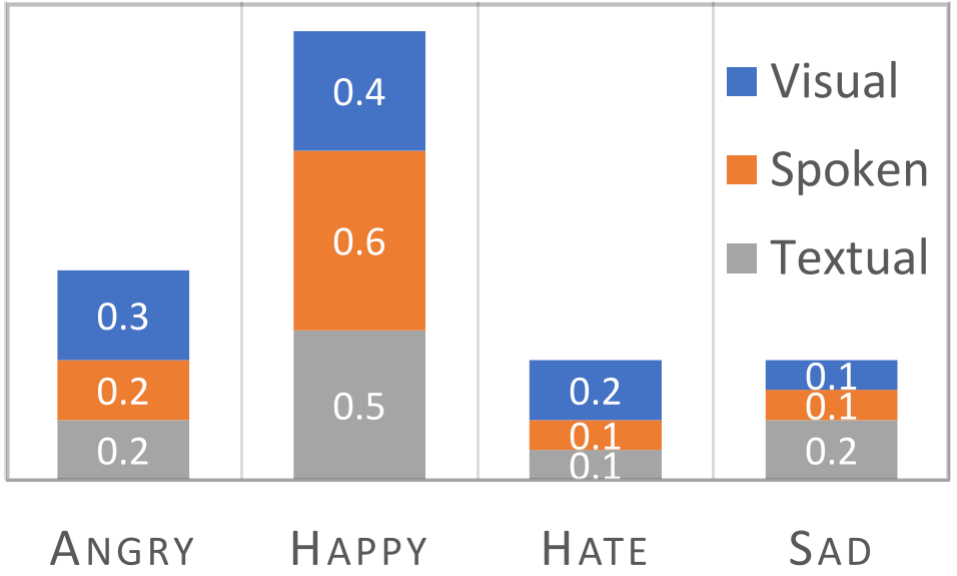} \vspace{-.05in}
\captionof{figure}{\centering Example of emotion label determination.\vspace{-.05in}}
\label{fig:label_formation}
\vspace{-.1in}
\end{figure}

\begin{itemize}
\item The text from the BT4SA dataset is pre-processed by removing links, special characters, and tags, and then the cleaned text is converted to speech using the pre-trained state-of-the-art text-to-speech (TTS) model, DeepSpeech3~\cite{ping2018deep}. The rationale for using TTS is based on recent studies showing that TTS models produce high-quality speech, which can serve as a reliable approximation of natural speech \cite{ping2018deep,oord2016wavenet}.\vspace{.01in} %deepmind2016wavenet,oord2016wavenet 

\item The image and speech components are passed through a pre-trained VGG model for IER and SER, while the text component is passed through a Bidirectional Encoder Representations from Transformers (BERT) model for TER. The VGG was trained on the Flickr \& Instagram (FI)~\cite{you2016building} dataset and the IEMOCAP~\cite{busso2008iemocap} dataset for IER and SER, respectively, while the BERT was trained on the ISEAR dataset \cite{scherer1994evidence} for TER. Prediction probabilities for each emotion class are obtained per modality. For recognition, we employed models distinct from those used in dataset construction, utilizing VGG (trained on ImageNet) for visual and speech modalities, and BERT (uncased L-12\_H-768\_A-12) for the textual modality.\vspace{.01in}
 
\item The prediction probabilities are averaged to obtain each sample’s ground‐truth emotion, ensuring that the chosen ground truth is the one that is supported by the majority of modalities. {Fig. \ref{fig:label_formation} shows an example of emotion label determination, whereas Table \ref{tab:data} describes the IIT-R MMEmoRec dataset's class-wise distribution. The probabilities for each emotion class given by each modality are shown. The `happy' class has an average prediction probability of $0.500$ compared to $0.233$ for `angry,' $0.133$ for `hate,' and $0.133$ for `sad.' The final emotion label for the sample is determined as `happy.'} \vspace{-.01in} %It is important to note that in b-t4SA, multiple images correspond to a single text and speech sample, so for such a sample, the text and speech are kept the same while repeating the same image for such cases hence giving us about 4.7M examples. 

\item The data is segregated according to classes, and the samples having an average prediction probability of less than the threshold confidence value of $0.55$ times the maximum probability for the corresponding class are discarded. The threshold confidence is determined in Section \ref{sec:ablation1}.\vspace{.03in}

\item The four emotion classes, `angry,' `happy,' `hate,' and `sad,' are common in various datasets of different modalities considered in this work. Samples labelled as `excitement' were merged with `happy,' as excitement, categorized under `surprise' in Plutchik’s wheel of emotions~\cite{plutchik2001nature}, shares its positive valence. Similarly, samples labelled as `disgust' were re-labelled as `hate,' aligning with their shared high arousal while `sad' denotes low arousal. The final dataset contains $112455$ samples with $53317$ labelled as `angry,' $44980$ as `happy,' and $10327$ \& $3831$ as `sad' and `hate' respectively. Table \ref{tab:sample_data} shows samples from the IIT-R MMEmoRec dataset.
\end{itemize}

\begin{table}[!t]
\centering	
\caption{Class-wise distribution of the data samples across each emotion class for Set A, Set B, and Overall (A+B).\vspace{-.05in}}
\label{tab:data}
\resizebox{.28\textwidth}{!}
{%
 \begin{tabular}{lccc}
    \toprule
    \textbf{Emotion} & \textbf{Set A} & \textbf{Set B} & \textbf{Overall}\\ \midrule
    Angry  & 53317    & 98947    & 153264   \\
    Happy  & 44980    & 97401    & 142381   \\
    Hate   &  3831    &  2313    &   6144   \\
    Sad    & 10327    &  1548    &  11875  \\ \hdashline
    Total  & \textbf{112455} & \textbf{200209} & \textbf{312664} \\\bottomrule
\end{tabular}%
}
\end{table}

% \setcounter{table}{0}
% \begin{minipage}{1\textwidth} 
%     \hspace{-.2in}
%     \begin{minipage}{0.18\textwidth}
%     \begin{center}
%     \captionof{table}{\centering Class-wise data distribution.\vspace{-.05in}}
%     \label{tab:data}
%     {\fontsize{8}{12}\selectfont
%          \begin{tabular}{@{}lc@{}}
%             \toprule
%             \textbf{Emotion}   & \textbf{Samples} \\ \midrule
%             Angry  & 53,317    \\
%             Happy  & 44,980    \\
%             Hate   &  3,831    \\
%             Sad    & 10,327    \\ \bottomrule
%         \end{tabular}%
%     }
%     \end{center}
%     \end{minipage} 
%     \hspace{-.05in}
%     \begin{minipage}{0.32\textwidth}
%     \begin{center}
%     %\renewcommand\thefigure{2}
%         \vspace{.1in}
%         \includegraphics[width=.9\textwidth]{fig_label_formation} \vspace{-.05in}
%         \captionof{figure}{\centering Example of emotion label determination.}
%         \label{fig:label_formation}
%     \end{center}
%      \end{minipage} 
%      %\hfill
% \end{minipage}\vspace{.08in}

%Actual Position: Elsewhere
\setcounter{table}{1}
\begin{table*}[!t]
\centering	
\caption{A few samples from IIT-R MMEmoRec dataset. Here, `Img\_Prob,' `Sp\_Prob,' `Txt\_Prob,' and `Final\_Prob' are image, speech, text, and final prediction probabilities, whereas angry, happy, hate, and sad emotion labels are denoted as 0, 1, 2 \& 3 respectively.\vspace{-.05in}}
\label{tab:sample_data}
\resizebox{.89\textwidth}{!}{%
    \begin{tabular}{l}
    \includegraphics[width=1\textwidth]{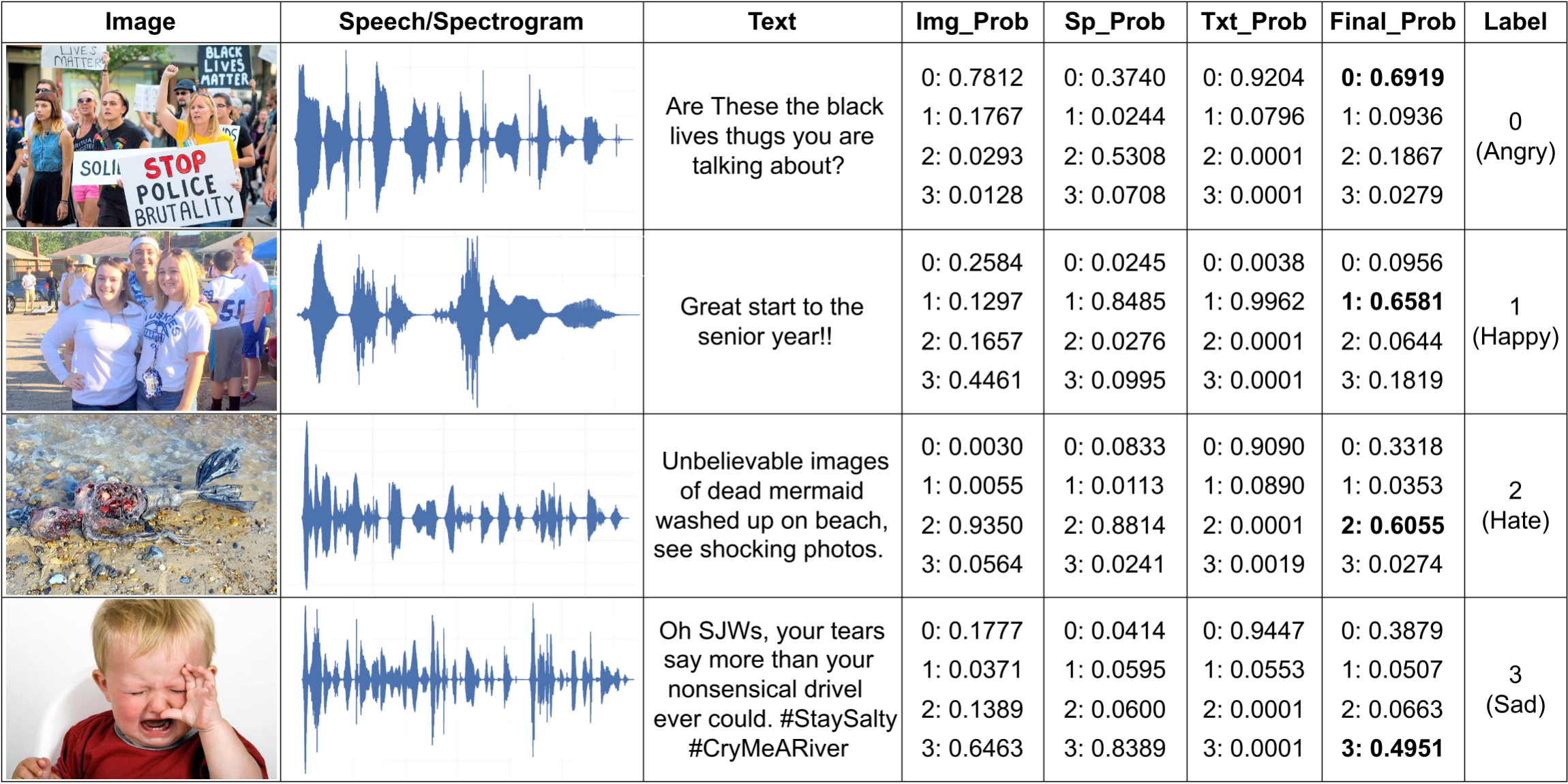}
    \end{tabular}%
} \vspace{-.1in}
\end{table*}  

% for Dataset Construction
\subsubsection{Determining Threshold Confidence Value} \label{sec:ablation1}
\begin{figure}[!h]
    \vspace{-.2in}
    \centering\hspace{-.15in}
    \subfloat[Number of data samples retained for different threshold values.]{%
        \includegraphics[width=0.49\textwidth]{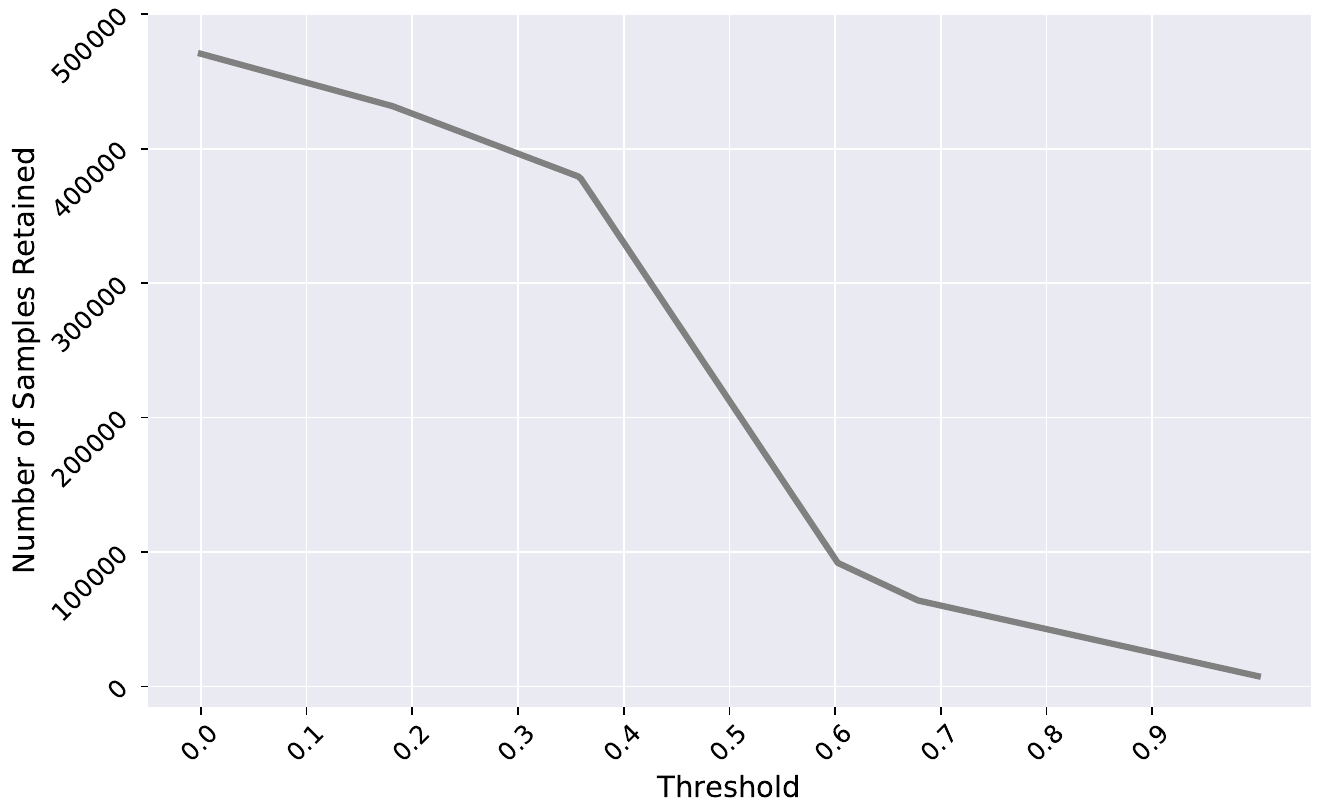}
        \label{fig:data_retained}
    }    
    
    \subfloat[Class-wise data distribution for different thresholds values.]{%
        \includegraphics[width=0.47\textwidth]{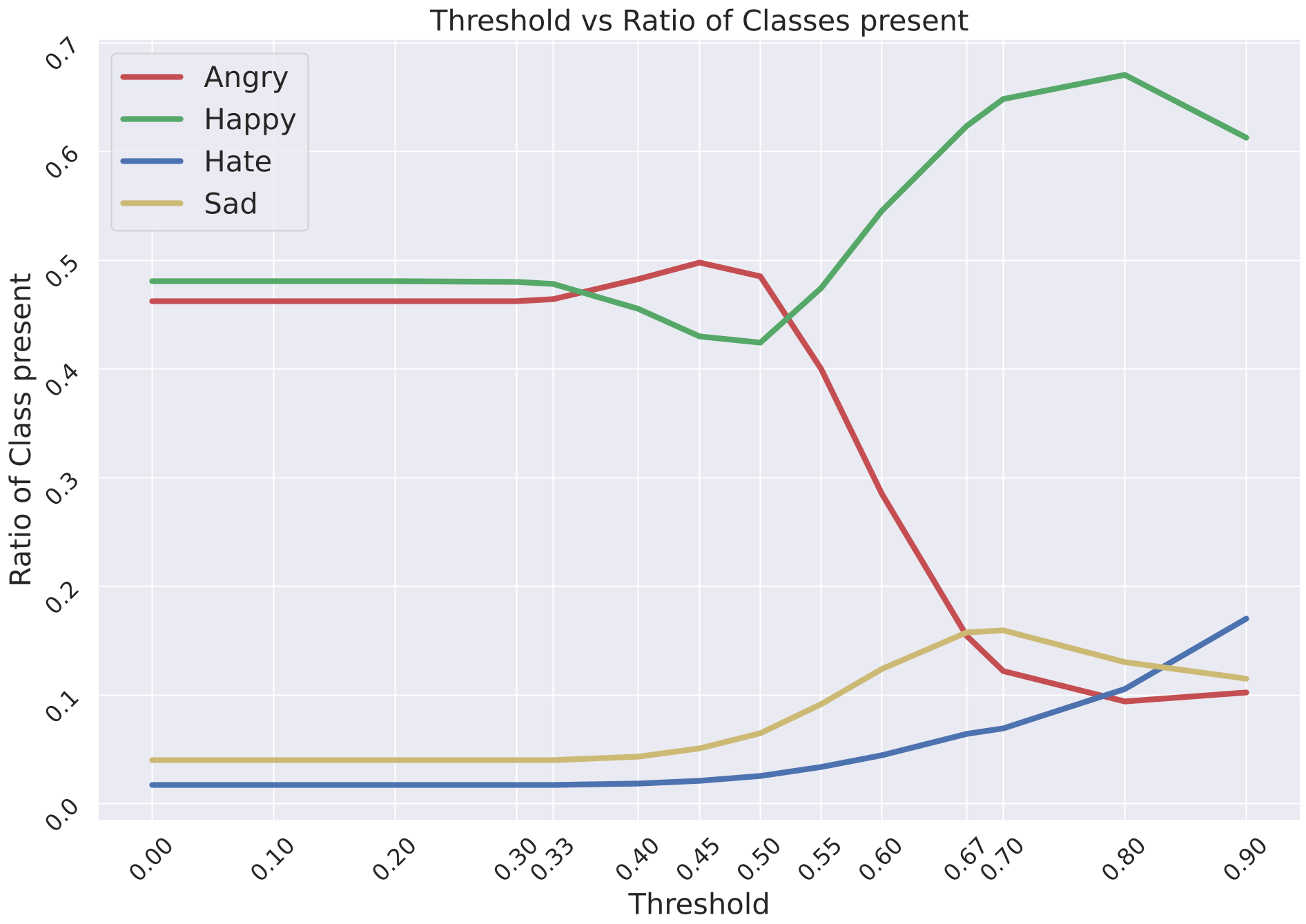}%
        \label{fig:class_distribution}
    }\vspace{-.07in}
    \caption{Determining threshold confidence for dataset construction.\vspace{-.2in}}
    \label{fig:data_combined}
\end{figure}

{The B-T4SA dataset comprises $470586$ samples labelled as `positive,' `negative,' and `neutral.' Samples labelled `neutral' were removed and the remaining $312664$ samples after manual cleaning were mapped to discrete emotion labels (`angry,' `happy,' `hate,' `sad') after processing the image, speech, and text components through respective emotion recognition models. This refined collection is referred to as Set A + Set B. Set A, containing $112455$ samples with a high degree of confidence in their emotion classification, is designated as the final IIT-R MMEmoRec dataset. Conversely, the additional $200209$ samples in Set B are with a lower degree of confidence. Both sets A and B of the dataset along with VISTANet's code can be accessed at \href{https://github.com/MIntelligence-Group/MMEmoRec}{\underline{github.com/MIntelligence-Group/MMEmoRec}}.} To establish the confidence threshold for Set A, these factors are considered:

\begin{itemize}
\item 
\textit{Modality-Wise Agreement}: The chosen threshold should ensure the final emotion class with the highest 
$final\_prob$ to have a minimum probability of $0.51$, affirming a decisive classification without inter-modality conflict. For instances, where one modality supports a secondary emotion class, the threshold ensures that the $final\_prob$ of this class remains below that of the primary emotion class supported by the other two modalities. It mandates a threshold greater than $0.51$. % for the corresponding sample to be retained in the IIT-R MMEmoRec dataset

\item 
\textit{Dataset Size}: Observations from Fig. \ref{fig:data_retained}, indicate that until a threshold of $0.37$, a significant number of samples are retained, suggesting a potential compromise in label confidence. Between thresholds of $0.37$ and $0.6$, there is a steep decline in sample retention, indicating a more stringent filtering of data quality. Above a threshold of $0.6$, very few samples are retained. Hence, a dataset between $0.37$ and $0.6$ should be selected to have a balance of dataset size and quality.

\item 
\textit{Class Distribution Consistency}: It is imperative to maintain a class distribution in the resultant dataset that mirrors that of the original B-T4SA dataset. Fig. \ref{fig:class_distribution} shows that maintaining the threshold up to $0.33$ preserves this distribution optimally, with thresholds up to $0.55$ still acceptable before the distribution significantly diverges.
\end{itemize}

Given these considerations, a threshold of $0.55$ emerges as the most effective choice, balancing high confidence in data labels, adequate sample retention, and class distribution's preservation.

% The original B-T4SA dataset contained 470586 data samples labelled as `positive,' `negative,' and `neutral.' While constructing the IIT-R MMEmoRec dataset with discrete emotion labels, i.e., `angry,' `happy,' `hate,' and `sad,' it was essential to retain only the samples having high confidence in the associated emotion label. After passing the image, speech, and text components of the inputs to respective emotion recognition models as discussed in Section \ref{sec:proposed}, we computed a value for each data sample in each class representing at what percentage compared to the class maximum that sample is in its particular class. It gave us the confidence of each data sample in its particular class. To determine the appropriate threshold, we plotted possible threshold values vs. the ratio of the class present (the number of each class sample and the total number of samples) as shown in Fig. \ref{fig:data}. The higher the threshold, the higher the confidence and the better the data quality. However, a higher threshold value also leads to two issues -- i) reduction in the size of the dataset and ii) disruption in the distribution of emotion classes compared to its original distribution. 

%An appropriate threshold value needs to be chosen, leading to a good trade-off between high confidence and the appropriate size \& distribution of the dataset. The distribution of each class at a threshold approaching $1$ is very different compared to when all samples are taken at the $0$ thresholds. Till the threshold value of $0.33$, the distribution is almost the same as the original, but this confidence is too low to be acceptable. The next possible value is above $0.5$ but below $0.6$. Between these two values, the distribution of various classes is almost the same, and the confidence is also above $0.5$, which is acceptable. Hence, an average value of $0.55$ is chosen as the threshold confidence value. 

\subsubsection{Human Evaluation}\label{sec:human_eval}
The IIT-R MMEmoRec dataset was evaluated by eight people. Two human readers (one male, one female) recorded the text components as speech, and evaluators compared these to the machine‐synthesized versions on a $0-100$ contextual‐similarity scale. They also rated how well each modality matched the annotated emotions. Average scores are shown in Table \ref{tab:human_eval}, where $S_{ss-hs}$ represents the percentage of evaluators who found the synthetic speech (ss) similar to human speech (hs). $S_{ss}$ \& $S_{hs}$ denote the percentages of synthetic and human speech components portraying the annotated emotion. $S_{i}$ and $S_{t}$ indicate the agreement of the image and text with the annotated emotion, respectively. $S_{ss-i-t}$ and $S_{hs-i-t}$ reflect the agreement across all three modalities considering synthetic and human speech.

\setcounter{table}{2}
\begin{table}[!h]
\centering	
\caption{IIT-R MMEmoRec dataset's human evaluation. Here $S_{ss-hs}$ denotes the similarity between synthetic speech (ss) and human speech (hs) reported by the evaluators. $S_{i}$, $S_{t}$, $S_{hs}$ and $S_{ss}$ denote the reported agreement of annotated emotion class with image, text, human speech and synthetic speech inputs respectively.\vspace{-.05in}}
\label{tab:human_eval}
\resizebox{.5\textwidth}{!}
{%
    \begin{tabular}{lccccccc}\toprule
    \textbf{Class} & $\mathbf{S_{ss-hs}}$ & $\mathbf{S_{ss}}$ & $\mathbf{S_{hs}}$ & $\mathbf{S_{i}}$ & $\mathbf{S_{t}}$ & $\mathbf{S_{ss-i-t}}$ & $\mathbf{S_{hs-i-t}}$ \\ \midrule
    {Angry}   \hspace{0in} & 67.18\% &82.81\% & 85.94\% & 70.31\% & 84.38\% &75.00\% & 82.03\% \\
    {Happy}   \hspace{0in} & 52.78\% &66.67\% & 69.44\% & 63.89\% & 72.22\% &66.67\% & 69.44\% \\
    {Hate}    \hspace{0in} & 62.50\% &71.43\% & 72.32\% & 67.86\% & 71.43\% &73.21\% & 72.32\% \\ 
    {Sad}     \hspace{0in} & 60.42\% &77.08\% & 78.13\% & 75.00\% & 87.50\% &77.08\% & 83.33\% \\ \hdashline
    {Overall} \hspace{0in} & \textbf{60.72}\% &\textbf{74.49}\% & \textbf{76.46}\% & \textbf{69.26}\% & \textbf{78.81}\% &\textbf{72.99}\% & \textbf{76.78}\% \\ \bottomrule
    \end{tabular}
}\vspace{-.07in}
\end{table}

We had two readers read the text of the data samples and called their output human-synthesized speech. 60.72\% of evaluators found the synthetic speech contextually similar to the human-synthesized speech. 74.49\% of synthetic speech samples and 78.91\% of human-synthesized speech samples were found to portray the annotated emotion labels. Additionally, 69.26\% of images and 78.81\% of text components of the data samples corresponded to the annotated emotion labels. Moreover, 72.99\% of the samples considering machine-synthesized speech with corresponding text \& image aligned with the determined emotion label, comparable to 76.74\% when considering human-synthesized speech with text \& image.

\subsubsection{Anthropomorphic Score based Evaluation}\label{sec:val_annot}
The Anthropomorphic Score, proposed by Jaiswal et al. \cite{jaiswal2019generative}, quantifies the human-like quality of synthesized speech. It represents the SER accuracy ratio between synthesized speech and real speech. We assessed the reliability of the IIT-R MMEmoRec dataset's speech component synthesized via TTS using this metric. To validate our approach, we tested it on two multimodal emotion recognition datasets: IEMOCAP \cite{busso2008iemocap} and MELD \cite{poria2018meld}, conducting SER using their real speech and speech synthesized from their text via DeepVoice3. The computed Anthropomorphic Scores for both datasets averaged $0.94$, confirming that TTS-generated speech reliably approximates real speech.

\subsection{VISTANet}\label{sec:proposed}  
\begin{figure*}[!h]
\centering
\includegraphics[width=1\textwidth]{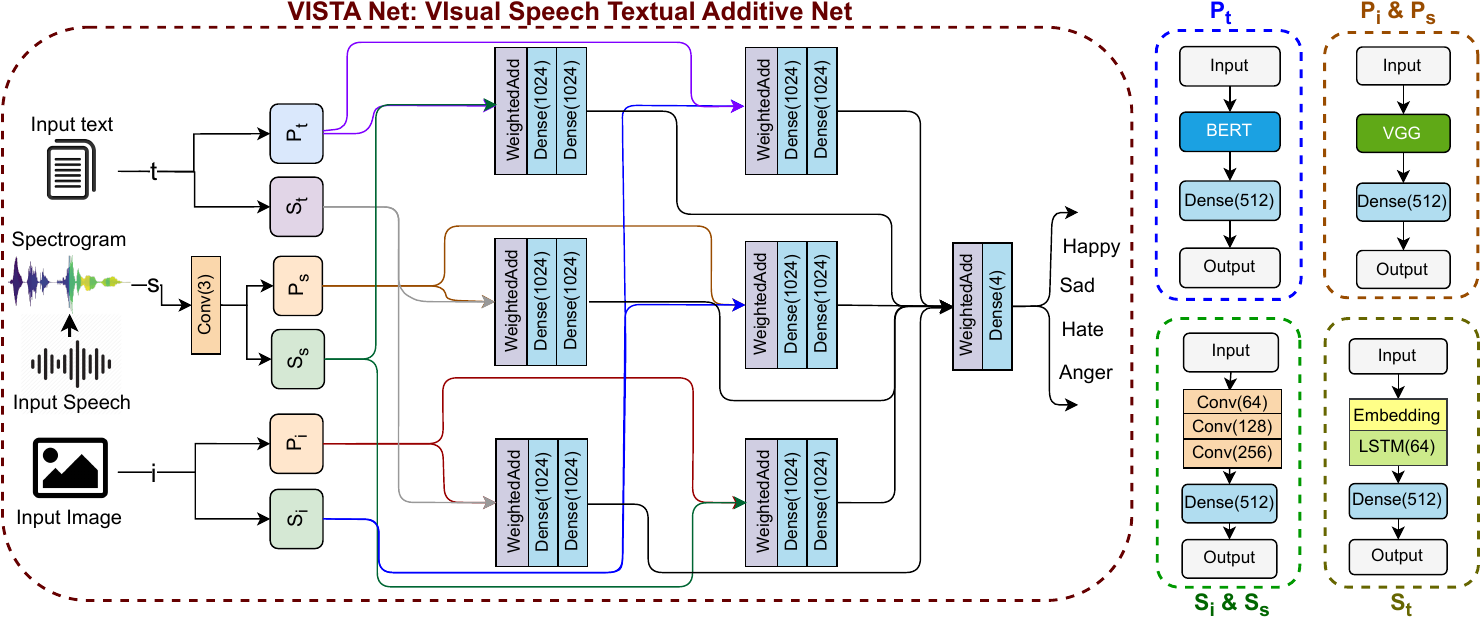}\vspace{-.1in} 
\caption{Schematic architecture of the proposed multimodal emotion recognition system. Here, $\mathbf{P_m}$ \& $\mathbf{S_m}$ denote the pre-trained \& simpler networks for $m^{th}$ modality whereas `i,' `s,' and `t' denote visual, speech and text modalities, respectively. The initial six blocks represent pairwise intermediate fusion and the final block illustrates the late fusion.\vspace{-.1in}}
\label{fig:archi}
\end{figure*}

The proposed system, VISTANet's architecture, is shown in Fig.~\ref{fig:archi}, which has been decided based on the ablation studies discussed in Section~\ref{sec:ablation2}. It fuses image, speech \& text features using a hybrid of two-stage intermediate (pairwise) and late fusion, which considers all possible pairs of all three modalities and automatically weights them without human intervention. {This staged approach combines the strengths of intermediate and late fusion. The intermediate fusion captures unique interactions between image, speech, and text modalities and then the model integrates these insights through late fusion for a comprehensive analysis.} 
%Intermediate fusion combines information from various modalities before classification, specifically after feature extraction, whereas late fusion combines information post-classification.

%Actual Position: elsewhere

The three modalities are fed into two types of networks: pre-trained and simpler networks. The intuition behind this approach is to build a fully automated multimodal emotion classifier by including various modalities in all possible combinations and learning their weights while training without any human intervention. The proposed system contains $P_i$ and $S_i$ for image, $P_s$ and $S_s$ for speech, and $P_t$ and $S_t$ for text, denoting pre-trained and simpler networks, respectively. The input speech has been converted to a log-mel spectrogram before being fed into the network. A combination of complex pre-trained models has been employed with simpler, adaptable models to enhance the system's efficiency and adaptability. This setup mirrors a dynamic where a structured, rule-following member (complex pre-trained model) provides robust foundational knowledge, guiding a flexible, adaptable member (simpler model). This arrangement allows the simpler models to adapt and apply these insights to new scenarios, thus maintaining the unique identity of each modality while optimizing overall system responsiveness. By leveraging the strengths of both model types, VISTANet ensures that learning is not only comprehensive but also sufficiently flexible, allowing for an effective integration of insights across different modalities.

The idea of the proposed method is inspired by emotion theories and related cognitive studies \cite{scherer1994evidence, russell1980circumplex} on how the human brain perceives emotions from complex environmental stimuli. The integration of complex pre-trained models with simpler models is designed to mimic the hierarchical and layered processing of emotions in the human brain \cite{adolphs2002neural, pessoa2008relationship}, effectively capturing both emotional contents and general (non-emotional) information as described in Appraisal Theories \cite{scherer1994evidence} and Dimensional Theories \cite{russell1980circumplex}. The human brain processes all information at the lower level neural pathways, including general information such as color, shape, pitch, etc., regardless of whether they are emotion-related or not. At a later phase, our brain integrates related attributes to form emotions or feelings. For example, a dark scene with a high-pitched sound could produce `fear’. Building on this foundation, VISTANet emulates how humans integrate multiple modalities, i.e., the pre-trained models perceive generic information, and the simpler but specific models focus on emotional attributes such as action units, speech tones, emotional words, etc., which are fused and their contributions are dynamically adjusted, mirroring the human tendency to prioritize certain channels over others depending on their informativeness for perceived emotions \cite{de2000perception}. 

\subsubsection{Intermediate Fusion Phase}\label{sec:if}
The images of dimension $(128, 128, 3)$ are resized to $(224, 224, 3)$ before being fed into $P_i$ and $S_i$ respectively. $P_i$ comprises a VGG16 model \cite{simonyan2014very} followed by a $512$-dimensional dense layer, while $S_i$ contains three convolution layers with $64$, $128$, and $256$ filters of size $(3, 3)$, followed by a dense layer of $512$ dimensions. The spectrogram of size $(128, 128, 1)$ from the speech input is initially processed through a convolution layer with $3$ filters of size $(3, 3)$ to enhance its feature extraction capability and to expand the channel depth to 3, making it compatible with the VGG16 model. This processed spectrogram is then further analyzed by $P_s$ and $S_s$, which consists of the architecture as $P_i$ and $S_i$ respectively.

The text input is processed by $T_i$, which includes a BERT model \cite{devlin2018bert}, and $T_s$, which consists of an embedding layer followed by an LSTM layer with $64$ units. Both $T_i$ and $T_s$ lead into $512$-dimensional dense layers. In the intermediate fusion step, all pairs of pre-trained and simpler networks from different modalities are combined using a `WeightedAdd' layer that we have defined. This results in six distinct combinations, each processed through two dense layers with $1024$ neurons, providing classification outcomes based on each pair. Equation \ref{eq:eq1} illustrates all possible pairings from the combination of pre-trained and simpler networks, ensuring that the networks in each pair do not belong to the same modality. {Pairwise fusion at the intermediate stage leverages the distinct strengths of each modality pair, enhancing the model's ability to capture nuanced intermodal interactions and improve emotion recognition accuracy before final integration.}\vspace{-.12in}

{\fontsize{9}{10}\selectfont 
	\begin{eqnarray}
	\label{eq:eq1}
	\begin{split}	
	%&y_{i}=softmax(concat(I_e,S_e)^{T}W+b)\\
	&O_1 = WeightedAdd (P_i,S_s)\\
	&O_2 = WeightedAdd (P_i,S_t)\\
	&O_3 = WeightedAdd (P_s,S_i)\\
	&O_4 = WeightedAdd (P_s,S_t)\\
	&O_5 = WeightedAdd (P_t,S_i)\\
	&O_6 = WeightedAdd (P_t,S_s)\\
	%&\begin{cases}
	%where:\\
	%F_m:\ 
	%\end{cases}\\
	\end{split}
	\end{eqnarray}
} 

\noindent where $O_1$, $O_2$, $O_3$, $O_4$, $O_5$, and $O_6$ represent the classification outputs for the pairs ($P_i$, $S_s$), ($P_i$, $S_t$), ($P_s$, $S_i$), ($P_s$, $S_t$), ($P_t$, $S_i$), and ($P_t$, $S_s$), respectively. The `WeightedAdd' layer ensures that during training, the weight of any weighted addition is learned using back-propagation without any human intervention. Each weight in the `WeightedAdd' layer is randomly initialized and passed from the softmax layer, giving us positive values used as final weights and learned during training.

\subsubsection{Late Fusion Phase}
In this phase, the information from various modalities' all possible pairs is combined in a hybrid manner. The intermediate classification outputs obtained from above Eq. \ref{eq:eq1} are passed from another `WeightedAdd' layer, which combines these outputs dynamically, giving us the final output $O$ as depicted in Eq. \ref{eq:eq2}. The output $O$ is passed from a dense layer with dimensions equal to the number of emotion classes, i.e., four.\vspace{-.2in}

\begin{align}
    \label{eq:eq2}
    &O = WeightedAdd(O_1,O_2,O_3,O_4,O_5,O_6)
\end{align}\vspace{-.15in}

\noindent where $O$ denotes the final output and $O_1$, $O_2$, $O_3$, $O_4$, $O_5$ and $O_6$ are the intermediate classification outputs. VISTANet dynamically learns the weight coefficients on a per-sample basis. Beyond averaging, it learns to capture the interactions among different modalities, which is crucial for a more nuanced understanding. %These capabilities are further validated by the model’s superior performance in sentiment classification (which relies on true labels) and emotion classification across other datasets such as IEMOCAP and MELD, both of which also utilize true labels [Tables 5 and 6]. This indicates that the model's learning transcends mere averaging, addressing the concerns raised about the paper's results.

\subsection{KAAP}\label{sec:inter_technique}
\begin{figure*}[!h]
\centering
\includegraphics[width=.92\textwidth]{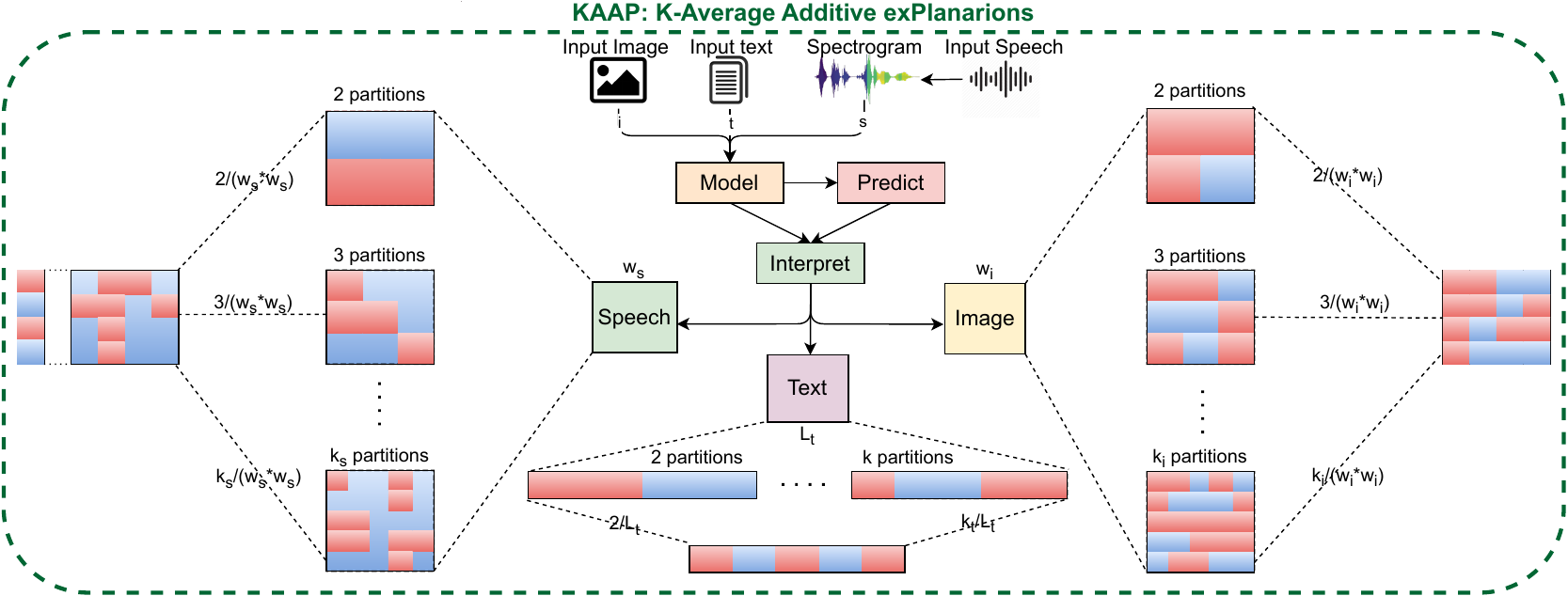} \vspace{-.07in}
\caption{Schematic representation of the proposed interpretability technique. The symbols $k_i$, $k_s$, and $k_t$ represent number of image, speech, and text partitions; $w_i$, and $w_s$ are the widths for image \& speech feature matrices, and $L_t$ is the length of the text feature vector.\vspace{-.1in}}  
\label{fig:inter}\vspace{-.05in}
\end{figure*}

This Section proposes a novel multimodal interpretability technique, K-Average Additive exPlanation (KAAP), depicted in Fig. \ref{fig:inter}. It computes the importance of each modality and its features while predicting a particular emotion class. Most of the existing interpretability techniques do not apply to speech and multimodal emotion recognition. Moreover, the most frequently used and accepted interpretability technique for images and text is Shapley Additive exPlanations (SHAP) \cite{lundberg2017unified}, which is an approximation of Shapley values \cite{SHAPley1953value}. It requires $O(n^2)$ computational time complexity, whereas KAAP requires a time of $O(k^2)$ where $k<=n$ is a given hyper-parameter. Moreover, KAAP applies to multimodal emotion analysis and a single modality or a combination of any two modalities.

\subsubsection{Calculating K-exPlanable Values}
\begin{figure}[!h]
	\centering
	\includegraphics[width=0.38\textwidth]{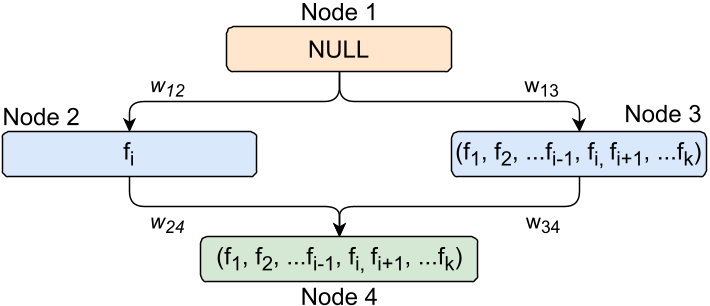} 
	\caption{Sample model for KP values computation.\vspace{-.2in}}
	\label{fig:kp_values} 
\end{figure}

\begin{spacing}{1.1}
For a model with $k$ features $\{f_1, f_2, \dots, f_k\}$, the K-exPlanable (KP) value of feature $f_i$, denoted $kp_{f_i}$, represents its importance. Fig. \ref{fig:kp_values} depicts an example calculation. Consider four nodes: Node 1 with no feature, i.e., NULL; Node 2 with a single feature $f_i$; Node 3 containing all remaining features from Node 1, i.e., $\{f_1, f_2, \dots, f_{(i-1)}, f_{(i+1)}, \dots, f_k\}$; and Node 4 with all the features $\{f_1, f_2, \dots, f_{(i-1)}, f_i, f_{(i+1)}, \dots, f_k\}$. The `Marginal Contribution' of an edge connecting Node $i$ and Node $j$ is defined as the difference between the prediction probabilities when using their respective features. For a given predicted label $c$, the marginal contribution of the feature $f_i$ from Node 1 to Node 2 is calculated as per Eq. \ref{eq:eq3}, where $prob_{\{f_i\}}$ is the probability of label $c$ calculated by using only feature $f_i$ and setting all other features to zero.\vspace{-.18in}
\end{spacing}

\begin{eqnarray}
	\label{eq:eq3}
	\begin{split}	
	&MC_{f_i,\{f_i\}} = prob_{\{f_i\}} - prob_{\{\phi\}}\\
	\end{split}
\end{eqnarray} \vspace{-.1in}

\noindent where $\phi$ denotes a null feature. The overall importance of $f_i$ is determined by calculating the weighted average of all `marginal contributions' of $f_i$ as shown in Eq. \ref{eq:eq4}.\vspace{-.15in}

\begin{align}
    \label{eq:eq4}
    KP_{\{f_i\}}(k) &= w_{12} \times MC_{f_i,\{f_i\}} \nonumber \\
    &\quad + w_{34} \times MC_{f_i,\{f_1, f_2, \dots, f_k\}}
\end{align}\vspace{-.1in}

\begin{spacing}{1.1}
The weights $w_{12}$ and $w_{34}$ must satisfy two conditions: i) their sum equals one to normalize the weights; ii) $w_{34}$ must be $(k-1)$ times $w_{12}$, reflecting the fact that $MC_{f_i, \{f_i\}}$ represents the contribution of adding $f_i$ to an empty set, while $MC_{f_i, \{f_1, f_2, \dots, f_k\}}$ considers its contribution to a nearly complete set of features. These relations are formulated in Eq. \ref{eq:eq5} and calculated as shown in Eq. \ref{eq:eq6}.  \vspace{-.12in}
\end{spacing}

\begin{align}
    \label{eq:eq5}
    &w_{12} + w_{34} = 1; \hspace{.2in} w_{12} = \frac{w_{34}}{k-1}
\end{align}\vspace{-.25in}

\begin{align}
    \label{eq:eq6}
    &w_{12} = \frac{1}{k}; \hspace{.2in} w_{34} = 1 - \frac{1}{k}
\end{align} \vspace{-.1in}

\noindent Eq. \ref{eq:eq7b} show the KP values calculated using Eqs. \ref{eq:eq4} and \ref{eq:eq6}.\vspace{-.1in}

\begin{align}
    \label{eq:eq7b}
    &KP_{\{f_i\}}(k) = \frac{1}{k} \times MC_{f_i,\{f_i\}} + (1 - \frac{1}{k}) \times MC_{f_i,\{f_1, f_2, \dots, f_k\}}
\end{align}\vspace{-.1in}

\subsubsection{Calculating KAAP Values}\label{sec:kaap_value}
The KAAP values are calculated to determine the importance of each modality and its features, enabling a detailed analysis of how individual and grouped elements contribute to emotion recognition. The information from image, text, and speech modalities is formatted continuously; for instance, a single pixel cannot alone define an object that may elicit a specific emotion in an image, but a collective array of pixels can. Similarly, while a spectrogram at a single instance of time \& frequency alone cannot define anything significant in speech, a meaningful segment containing multiple instances can. Likewise, in text, a single letter may not convey emotion, but a word composed of multiple letters certainly can. KAAP values, based on this nuanced understanding, are derived using the KP values for various feature groups within each modality.

First, the {input of size $l$} is divided into $k$ parts, where $k$ is a hyperparameter decided through the ablation study in Section \ref{sec:ablation}. These $k$ parts correspond to the $k$ features of the input. Then, for a feature group $f_i$, $KP_{f_i}(k)$ values are computed for the given value of $k$ using Eq \ref{eq:eq6}. It represents how a group of features $f_i$ will perform compared to all remaining groups. However, these groups can vary in size, i.e., $k$ can have various values that lead to different groups and thus to different KP values from groups of different sizes, thus affecting the original features' importance. To deal with this issue, the weighted average of all the KP values is taken in Eq. \ref{eq:eq7} for {$j \in \{2, 3 \dotsc, k\}$} where weights are equal to the number of features in that group of features. $k$ = 1 is ignored here as the whole input as one feature will not make any sense.\vspace{-.2in}

\begin{align}
    \label{eq:eq7}
    &kaap_{\{f_i\}} = \sum_{j=2}^{k} [\frac{j}{l}\times KP_{\{f_i\}}(j)] \hfill \triangleright {\color{gray}for\ linear\ feature}
\end{align}  

For input image and speech spectrogram, both of width $128$ and height $128$, their KP values for a given $k$ are calculated by dividing the input into $k$ parts along both axes. As a matrix defines image and speech spectrogram, this gives us a $k*k$ feature group. The equation for calculating the KAAP values for the above two inputs is given by  Eq. \ref{eq:eq8} {where $l$ denotes the input length for text and $w$ denotes the input width of the image or spectrogram feature matrices, each serving as a fixed normalization bound}. It gives us a matrix showing the importance of each pixel for a given image and speech input. This matrix directly represents the importance of the image. At the same time, for speech input, the values are averaged along the frequency axis to reduce the KAAP value matrix to the time axis, hence giving importance to speech at a given time.\vspace{-.2in}

\begin{align}
    \label{eq:eq8}
    &kaap_{\{f_i\}} = \sum_{j=2}^{k} [\frac{j^2}{w^2}\times KP_{\{f_i\}}(j)] \hfill \triangleright \hspace{-.03in} {\footnotesize \color{gray}for\ matrix\ feature \hspace{-.06in}}
\end{align}  

For input text, the division is done such that each text word is considered a feature, as the emotion can only be defined by a word, not a single letter, as discussed above. Then, the text is divided into $k$ parts, and as a linear array can represent text, the KAAP values are calculated using Eq. \ref{eq:eq7}. Also, the value of $k$ used for image, speech, and text modalities have been determined as $7$, $7$, and $5$, respectively, in Section \ref{sec:ablation3}. Furthermore, the modalities' importance defined by symbols $\upsilon$, $\delta$, and $\tau$ for visual, spoken, and textual features, respectively, are computed assuming that image, speech, and text are three distinct features and calculating each modality's KAAP value for $k$ = $3$. While evaluating one modality, the others are perturbed to zero.%The KAAP technique is depicted in Algorithm \ref{algo:a3}, which uses Algorithm \ref{algo:a2} that calculates the KAAP values for each data instance and Algorithm \ref{algo:a1} for probability prediction.

%-*-*-*-*-*- Implementation -*-*-*-*-*-*
%=======================================	
\section{Implementation}\label{sec:experiments}
\subsection{Experimental Setup}
The network training for the proposed system has been carried out on Nvidia Tesla V100 GPU, whereas the testing \& evaluation have been done on an Intel(R) Core(TM) i7-8700 Ubuntu machine with 64-bit OS and 3.70 GHz, 16GB RAM.

\subsection{Training Strategy and Hyperparameter Setting}\label{sec:dataset}
The model training has been performed using a batch-size of $64$, with data partitioned into training, validation, and testing sets at ratios of $70\%$, $15\%$, and $15\%$, respectively, and evaluated using 5-fold cross-validation, \textit{Adam} optimizer, \textit{ReLU} activation function with a learning rate of $1 \times 10^{-4}$ and \textit{ReduceLROnPlateau} learning rate scheduler with a patience value of $2$. The baselines and proposed models converged in terms of validation loss in $10$ to $15$ epochs. As a safe upper bound, the models have been trained for $50$ epochs with \textit{EarlyStopping} \cite{prechelt1998early} with patience values of $5$. The loss function is the average of categorical focal loss~\cite{lin2017focal} and categorical cross-entropy loss. Accuracy and \textit{CohenKappa} \cite{vieira2010cohen} have been analyzed for the model evaluation. %Macor f1 \cite{opitz2019macro}

%%%%%%%%
\subsection{Baselines and Proposed Models}\label{sec:models}
The `Image + Speech + Text' configuration described in Section \ref{sec:ablation_archi} is taken as Baseline 1. In Baselines 2–6, this simple fusion is replaced by each of the six pairwise fusion schemes defined in Eq.~\ref{eq:eq9}, allowing us to isolate and evaluate the contribution of every modality pair. All baselines share the same feature‐extraction backbones, classification head, and employ KAAP for interpretability and are made on a common idea, as described below. Firstly, all three modalities are fed into  $P_i$, $P_s$, $S_i$, $S_s$, $T_i$ and $T_s$ as described in Section \ref{sec:proposed}, and are then passed from a dense layer of $512$ neurons, resulting in a $512$-dimensional outputs which are then combined using `WeightedAdd' to give three outputs. The following strategy is being followed for combining them: any pre-trained network must be combined with another simpler network. At least one combination must contain the network from different modalities because if all the modalities combine with themselves, then such a combination will not lead to any information exchange. Thus, six such configurations are possible, as described in Eq. \ref{eq:eq9}.\vspace{-.12in}

{\fontsize{9}{10}\selectfont 
	\begin{eqnarray}
	\label{eq:eq9}
	\begin{split}	
	%&y_{i}=softmax(concat(I_e,S_e)^{T}W+b)\\
	&(\#1) : {\{P_i+S_i, P_s+S_s, P_t+S_t}\} \\
	&(\#2) : {\{P_i+S_i, P_s+S_t, P_t+S_s}\} \\
	&(\#3) : {\{P_i+S_s, P_s+S_i, P_t+S_t}\} \\
	&(\#4) : {\{P_i+S_s, P_s+S_t, P_t+S_i}\} \\
	&(\#5) : {\{P_i+S_t, P_s+S_i, P_t+S_s}\} \\
	&(\#6) : {\{P_i+S_t, P_s+S_s, P_t+S_i}\} \\
	%&\begin{cases}
	%where:\\
	%F_m:\ 
	%\end{cases}\\
	\end{split}
	\end{eqnarray}
} 

Configuration $(\#1)$ is discarded because it fails to meet the requirement that at least one combination must include different modalities. Configurations $(\#2)$, $(\#3)$, and $(\#6)$ are \textit{partially-complete} as they involve networks from the same modalities, while $(\#4)$ and $(\#5)$ are \textit{complete}. This strategy reveals two issues: i) only two out of five baselines are complete; ii) different datasets require different modal strengths. To adapt to various datasets and scenarios, the VISTANet system is introduced. It merges the outputs of baselines $2$-$6$, avoids self-combinations, and dynamically adjusts weights to meet dataset and scenario needs. 

% The configuration $(\#1)$ is discarded as it does not hold the condition that at least one combination must combine with a different modality. The configurations $(\#2)$, $(\#3)$, $(\#6)$ are \textit{partially-complete} combinations as one of the three outputs of these combinations combine the pre-trained and simpler network from the same modalities. On the other hand, the configurations $(\#4)$ and $(\#5)$ are \textit{complete}. Using this strategy presents two disadvantages: i) only two out of five such baselines are complete, with others partially complete; ii) different datasets have varying requirements. For instance, some multimodal datasets may have superior image and speech components, while others excel in text quality. To address any dataset and scenario, the automated multimodal emotion recognition system VISTANet has been proposed. It combines the outputs of baselines 2-6, excluding any self-combinations and taking the weighted average of the rest. Thus, it automatically adjusts the weights of each combination based on the problem statements and dataset requirements. The results for the baselines and the proposed system are summarized in the following section.

\begin{table*}[!b]
\centering
\caption{{Emotion recognition results on Set A, Set B, and overall for the IIT-R MMEmoRec dataset. Baseline 1 uses simple image + speech + text (IST) fusion while baselines 2–6 use the pairwise fusion configurations from Eq.~\ref{eq:eq9}. Here, `Acc,' `F1,' `CK,' `P,' `R' denote accuracy, F1-score, CohenKappa score, precision, and recall. The $\uparrow$ indicates that their higher values are better while the highest value is in bold. `T' denotes average per-sample interpretability inference time in seconds. The $\downarrow$ indicates that its lower values are better.} \vspace{-.05in}}
\label{tab:res_added}
\resizebox{1\textwidth}{!}{%
    \begin{tabular}{@{}lcccccc@{\hspace{.1in}}ccccccc@{\hspace{.1in}}ccccccc@{}}
    \toprule
    \multirow{2}{*}{\textbf{Model}} & \multicolumn{6}{c}{\textbf{Set A}} & \multicolumn{7}{c}{\textbf{Set B}} & \multicolumn{7}{c}{\textbf{Overall}} \\ 
    \cmidrule(r{.1in}){2-7} \cmidrule(lr{.1in}){9-14} \cmidrule(l){16-21}
    & \textbf{Acc}$\uparrow$ & \textbf{F1}$\uparrow$ & \textbf{CK}$\uparrow$ & \textbf{P}$\uparrow$ & \textbf{R}$\uparrow$ & \textbf{T}$\downarrow$ 
    &     & \textbf{Acc}$\uparrow$ & \textbf{F1}$\uparrow$ & \textbf{CK}$\uparrow$ & \textbf{P}$\uparrow$ & \textbf{R}$\uparrow$ & \textbf{T}$\downarrow$ 
    &     & \textbf{Acc}$\uparrow$ & \textbf{F1}$\uparrow$ & \textbf{CK}$\uparrow$ & \textbf{P}$\uparrow$ & \textbf{R}$\uparrow$ & \textbf{T}$\downarrow$  \\ 
    \midrule
    Baseline 1 (IST + KAAP) &86.60\% & 0.87 & 0.78 & 0.86 & 0.87 & 26.81 
                    &&67.82\% & 0.51 & 0.55 & 0.50 & 0.52 & 26.94 
                    &&72.29\% & 0.60 & 0.60 & 0.59 & 0.60 & 26.62 \\ 
    Baseline 2 (\#2 + KAAP) &94.89\% & 0.95 & 0.91 & 0.95 & 0.95 & 26.58 
                    &&73.96\% & 0.67 & 0.61 & 0.67 & 0.67 & 26.72 
                    &&78.94\% & 0.74 & 0.68 & 0.74 & 0.74 & 26.97 \\ 
    Baseline 3 (\#3 + KAAP) &95.44\% & 0.94 & 0.92 & 0.93 & 0.95 & 26.76 
                    &&74.43\% & 0.68 & 0.62 & 0.67 & 0.68 & 26.89 
                    &&79.43\% & 0.74 & 0.69 & 0.73 & 0.74 & 27.07 \\ 
    Baseline 4 (\#4 + KAAP) &95.39\% & 0.95 & 0.93 & 0.95 & 0.95 & 26.63 
                    &&74.92\% & 0.68 & 0.62 & 0.68 & 0.67 & 26.95 
                    &&79.79\% & 0.74 & 0.69 & 0.74 & 0.74 & 26.85 \\ 
    Baseline 5 (\#5 + KAAP) &95.58\% & 0.96 & 0.93 & 0.96 & 0.96 & 26.90 
                    &&74.71\% & 0.71 & 0.63 & 0.70 & 0.71 & 27.01 
                    &&79.69\% & 0.77 & 0.70 & 0.76 & 0.77 & 26.70 \\ 
    Baseline 6 (\#6 + KAAP) &95.37\% & 0.95 & 0.92 & 0.95 & 0.95 & 26.84 
                    &&74.87\% & 0.71 & 0.63 & 0.71 & 0.71 & 26.67 
                    &&79.75\% & 0.77 & 0.70 & 0.77 & 0.77 & 26.99 \\ 
                    \hdashline
    VISTANet + SHAP
    & \multirow{2}{*}{\textbf{95.99}\%} 
      & \multirow{2}{*}{\textbf{0.96}} 
      & \multirow{2}{*}{\textbf{0.93}} 
      & \multirow{2}{*}{\textbf{0.96}} 
      & \multirow{2}{*}{\textbf{0.96}} 
      & 43.89 
    &  & \multirow{2}{*}{\textbf{75.13}\%} 
      & \multirow{2}{*}{\textbf{0.72}} 
      & \multirow{2}{*}{\textbf{0.67}} 
      & \multirow{2}{*}{\textbf{0.72}} 
      & \multirow{2}{*}{\textbf{0.72}} 
      & 43.73 
    &  & \multirow{2}{*}{\textbf{80.11}\%} 
      & \multirow{2}{*}{\textbf{0.78}} 
      & \multirow{2}{*}{\textbf{0.73}} 
      & \multirow{2}{*}{\textbf{0.78}} 
      & \multirow{2}{*}{\textbf{0.78}} & 43.68 \\
    VISTANet + KAAP
    &  &  &  &  &  & 26.86 
    &&  &  &  &  &  & 26.81 
    &&  &  &  &  &  & 26.78 \\
    \bottomrule
    % VISTANet + SHAP &95.99\% & 0.96 & 0.93 & 0.96 & 0.96 & 43.89 
    %                 && 75.13\% & 0.72 & 0.67 & 0.72 & 0.72 & 43.73 
    %                 &&80.11\% & 0.78 & 0.73 & 0.78 & 0.78 & 43.68 \\
    %                 %&95.83\% & 0.95 & 0.92 & 0.94 & 0.95 & 43.89 
    %                 %&&75.00\% & 0.72 & 0.65 & 0.72 & 0.72 & 43.73 
    %                 %&&79.97\% & 0.77 & 0.71 & 0.77 & 0.77 & 43.68 \\ 
    % VISTANet + KAAP &\textbf{95.99}\% & \textbf{0.96} & \textbf{0.93} & \textbf{0.96} & \textbf{0.96} & \textbf{26.86} 
    %                 &&\textbf{75.13}\% & \textbf{0.72} & \textbf{0.67} & \textbf{0.72} & \textbf{0.72} & \textbf{26.81} 
    %                 &&\textbf{80.11}\% & \textbf{0.78} & \textbf{0.73} & \textbf{0.78} & \textbf{0.78} & \textbf{26.78} \\
    % \bottomrule
    \end{tabular}
}\vspace{-.05in}
\end{table*} 

%================================
%-*-*-*-*-*- Results -*-*-*-*-*-*
%================================    
\section{Results and Discussion}\label{sec:results} 
%The emotion classification results have been discussed in this Section, along with their interpretation and a comparison with the results of the existing methods. 

\subsection{Quantitative Results}\label{sec:acc}
{The VISTANet has achieved emotion recognition accuracies of 95.99\%, 75.13\% and 80.11\% for Set A, Set B and the overall dataset described in Section \ref{sec:ablation1}}. The detailed results, along with the results of various baseline models, are shown in Table \ref{tab:res_added}. {VISTANet and baseline models employ KAAP for interpretability, yielding an average inference time of approximately $26.8$ seconds per sample. Replacing KAAP with SHAP in VISTANet increased the inference time to $43.9$ seconds i.e. a $1.63$ times slow-down while keeping the classification metrics (Acc, F1, CK, P, R) unchanged. Similar slowdown was observed for the baseline models on replacing KAAP with SHAP for interpretability.}
%class-wise accuracies are shown in Fig.~\ref{fig:confmat} while its 

% \begin{figure}[!h]
% %\renewcommand\thefigure{7}
% \centering
% \includegraphics[width=0.33\textwidth]{fig_Confusion_matrix}
% \caption{Confusion matrix showing class-wise accuracies.} 
% \label{fig:confmat}
% \end{figure}  

%Original Position: \subsubsection{Sample Predictions}
\begin{figure*}[!h]
\centering 
\includegraphics[width=.95\textwidth]{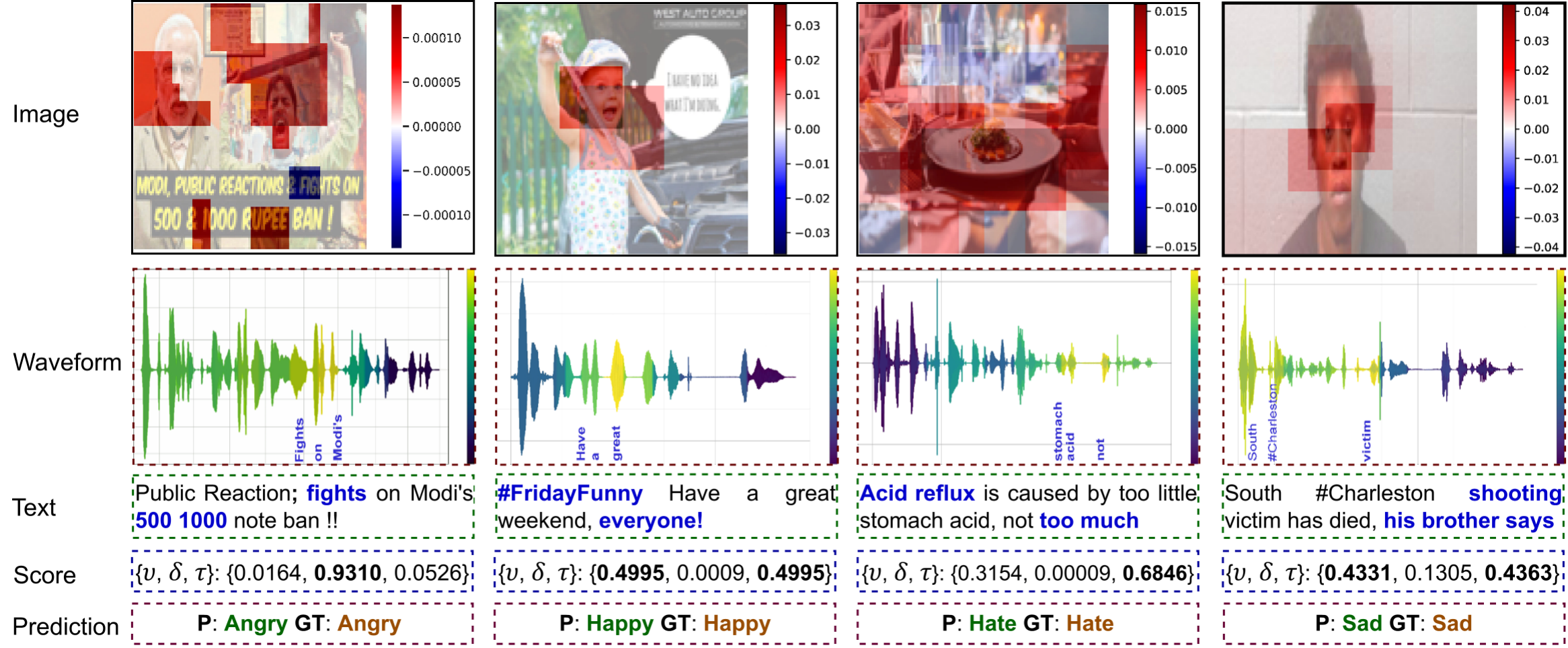}\vspace{-.05in}
\caption{Sample results. `P' \& `GT:' predicted \& ground-truth labels. `Score:' visual ($\upsilon$), speech ($\delta$) \& textual ($\tau$) modalities' importance.}
\label{fig:sample_res}
\end{figure*}
 
\subsection{Qualitative Results}\label{sec:qualres}	    
Fig.~\ref{fig:sample_res} shows sample emotion classification \& interpretation results. The key speech and image features for emotion classification are identified, with corresponding words highlighted. In the waveform, yellow and blue indicate the most and least important features, respectively. Speech and text are observed to be the most contributing to predicting the `angry' and `hate' classes, while image and text equally influence the `happy' and `sad' classes.%{These results show contributions for individual samples: voice and text dominate in the anger example and image and text dominate in the excitement example, while modality importance may vary across other samples.}

\subsection{Results Comparison}
\subsubsection{Emotion Recognition Results' Comparison}
Table \ref{tab:EmoRecExp} summarizes the emotion recognition results on the IIT-R MMEmoRec, IEMOCAP \cite{busso2008iemocap}, and MELD \cite{poria2018meld} datasets for VISTANet and state-of-the-art (SOTA) multimodal emotion recognition methods. It is important to note that VISTANet processes images for visual modality input, in contrast to some SOTA methods that utilize video inputs. To adapt SOTA methods for the IIT-R MMEmoRec dataset, we replicated the same image to match the frame requirements of different methods. For instance, while MER-MULTI uses the average of all frames, we utilized the original image features directly. Conversely, the Multimedia Information Bottleneck (MIB) method aligns frames with the number of words in the text, requiring us to copy the image features as many times as there are words. {Self-distillation (SDT) for emotion recognition in conversations (ERC)~\cite{ma2023transformer} and graph neural network (GNN) for ERC~\cite{chen2023multivariate} focus on processing conversations by considering past emotions to predict current ones, in contrast to our work on multimodal emotion recognition which predicts each data instance's emotion independently. We implemented these methods on the IEMOCAP and MELD datasets, maintaining the same conditions aimed at classifying emotions independently of the previous context. However, the lack of documented preprocessing steps for these datasets prevented us from preparing the MMEmoRec dataset in the necessary format, thus, SDT and GNN methods were applied only to IEMOCAP and MELD.}

\begin{table}[!h]
\centering
\caption{{Comparison of accuracies for emotion classification on the IIT-R MMEmoRec, IEMOCAP, and MELD datasets. The best and second-best results are marked in bold and underlined.\vspace{-.05in}}}
\label{tab:EmoRecExp}
\resizebox{.5\textwidth}{!}{% 
    \begin{tabular}{lcccccc}
    \toprule
    \textbf{Approach} & \multicolumn{3}{c}{\textbf{MMEmoRec}} & \textbf{IEMOCAP} & \textbf{MELD} \\ \cmidrule(lr){2-4}
                      & \textbf{Set A} & \textbf{Set B} & \textbf{Overall} & & \\
    \midrule
    DialogueGCN~\cite{ghosal2019dialoguegcn}    & -- & -- & -- 
                                                & 65.25\% & 59.46\% \\
    MEmoBERT~\cite{zhao2022memobert}            & 90.56\% & 70.29\% & 75.12\% 
                                                & 65.74\% & --\\
    MER~\cite{lian2023mer}                      & 92.17\% & 73.15\% & 77.68\% 
                                                & 66.13\% & 61.97\% \\
    DialogueCRN~\cite{hu2021dialoguecrn}        & -- & -- & -- 
                                                & 66.05\% & 60.73\% \\
    MM-DFN~\cite{hu2022mm}                      & -- & -- & -- 
                                                & 68.21\% & 62.49\% \\
    MIB~\cite{mai2022multimodal}                & 93.12\% & 73.67\% & 78.30\% 
                                                & 68.64\% & -- \\
    SDT for ERC~\cite{ma2023transformer}        & -- & -- & -- 
                                                & 67.93\% & 64.25\% \\
    GNN for ERC~\cite{chen2023multivariate}     & -- & -- & -- 
                                                & 69.26\% & 64.71\% \\
    UniMSE~\cite{hu2022unimse}                  & \underline{94.23}\% 
                                                & \underline{74.09}\% 
                                                & \underline{78.89}\% 
                                                & \textbf{70.56}\% 
                                                & \underline{65.09}\% \\
                                                \hdashline
    VISTANet                                    & \textbf{95.99}\% 
                                                & \textbf{75.13}\% 
                                                & \textbf{80.11}\% 
                                                & \underline{69.42}\% 
                                                & \textbf{65.20}\% \\
    \bottomrule 
    \end{tabular}
}\vspace{-.1in}
\end{table}

As observed from Table \ref{tab:EmoRecExp}, VISTANet either outperforms or closely competes with the SOTA methods across all datasets. This underscores its capability to handle diverse emotion recognition scenarios effectively. Additionally, the successful application of SOTA emotion recognition methods on the IIT-R MMEmoRec dataset further validates its reliability and usefulness. Furthermore, the experiments conducted on the IIT-R MMEmoRec dataset are speaker-dependent, as all speech samples were generated using the TTS strategy described in Section \ref{sec:proposed}. In contrast, the experiments on the IEMOCAP and MELD datasets are speaker-independent. 

\subsubsection{Sentiment Classification Results' Comparison}
The IIT-R MMEmoRec dataset has been constructed from the B-T4SA dataset in this paper; hence, there are no existing emotion recognition results for it. However, sentiment classification (into `neutral,' `negative,' and `positive' classes) results on the B-T4SA dataset are available in the literature, which have been compared with VISTANet's sentiment classification results in Table~\ref{tab:sota}.%{\color{blue} Incremental work \dots Explains why comparing SoTA results on different modalities, R2: no existing work on speech. This work performs better, although it does not incorporate extra information in speech such as voice tone and word stress} %"Moreover, this work has surpassed our previous work [10], 
% We have also compared the results for sentiment classification on the B-T4SA dataset and found the proposed model to perform better than the state-of-the-art (SOTA) methods.
% {\color{blue} [ToDo] 
% a) To run our model for sentiment analysis on full data (rather than the one prepared for emotion analysis)\\
% b) Specify the modalities (T+I, T+S, T+I+S) in the table\\
% }

\begin{table}[!h]
\centering
{%\fontsize{8}{11}\selectfont
\caption{Results comparison for sentiment classification on BT4SA dataset with existing approaches. Here, `V,' `S,' and `T' denote visual, spoken and textual modalities.\vspace{-.05in}}
\label{tab:sota}
\resizebox{.46\textwidth}{!}
{%
    \begin{tabular}{@{}lcc@{}}
        \toprule
        \textbf{Approach} & \textbf{Modality}   & \multicolumn{1}{c}{\textbf{Accuracy}} \\ \midrule
        Cross-Modal Learning~\cite{Vadicamo_2017_ICCVW} & V + T & 51.30\% \\
        Multimodal Sentiment Analysis~\cite{gaspar2019multimodal} & V + T & 60.42\% \\
        Hybrid Fusion~\cite{kumar2021hybrid} & V + T & 86.70\% \\
        Automated ML~\cite{lopes2021automl} & V + T  & 95.19\% \\
        \hdashline
        VISTANet   & \hspace{.03in} V + S + T \hspace{.03in} & \textbf{96.59}\% \\ \bottomrule 
    \end{tabular}
    }
}\vspace{-.1in}
\end{table}

\subsection{Ablation Studies}\label{sec:ablation}
The ablation studies have been performed {on set A} to determine the threshold confidence for data construction, appropriate network configuration for VISTANet, and suitable $k$ values for KAAP.

\subsubsection{Ablation Study 1: Determining Baselines and Proposed System's Architecture} \label{sec:ablation2}\label{sec:ablation_archi}
To begin with, the emotion recognition has been performed for a single modality at a time, i.e., separate IER, SER, and TER using pre-trained VGG models \cite{simonyan2014very} for Image \& speech and BERT \cite{devlin2018bert} for text. The performance has been evaluated in terms of Accuracy, CohenKappa metric (CK), F1 score, Precision, and Recall and summarized in Table \ref{tab:res}. The CK metric measures whether the distribution of the predicted class is in line with the ground truth.  

\begin{table}[!h]
\centering	
\caption{Ablation Study 1. Here, `Acc,' `F1,' `CK,' `P,' and `R' denote accuracy, F1-score, CohenKappa score, precision and recall. The highest values are marked in bold.\vspace{-.05in}}
\label{tab:res}
\resizebox{.43\textwidth}{!}{%
    \begin{tabular}{@{}lcccccc@{}}
    \toprule
    \textbf{Model} & \textbf{Acc} & \textbf{F1} & \textbf{CK} & \textbf{P} & \textbf{R} \\ \midrule
    Image only & 60.44 & 0.60 & 0.32 & 0.60 & 0.60\\
    Speech only & 78.69  & 0.75 & 0.62 & 0.74 & 0.79 \\
    Text only & 81.51  & 0.81 & 0.69 & 0.81 & 0.82 \\
    Image + Text & 86.40 & 0.86 & 0.77 & 0.86 & 0.86\\
    Image + Speech & 84.66 & 0.85 & 0.77 & 0.85 & 0.85\\
    Text + Speech & 81.95 & 0.81 & 0.70 & 0.82 & 0.81\\ \hdashline
    Image + Speech + Text & \textbf{86.60} & \textbf{0.86} & \textbf{0.78} & \textbf{0.86} & \textbf{0.87}\\ \bottomrule
    \end{tabular}
}\vspace{.05in} 
\end{table}

Next, we move on to the combination of two modalities. The chosen two modalities are fed into respective pre-trained models and then passed from a dense layer with $512$ neurons. Then the information from these modalities is added using the `WeightedAdd' layer defined in \ref{sec:if}. This output is next passed from three dense layers of size $1024$, $1024$, and $4$ neurons, which then classifies the emotion. An additional evaluation confirmed that combining two modalities outperforms individual modalities.

Finally, the information from all three modalities is combined and fed into their respective pre-trained models and is then passed from a dense layer of size $512$, which is then passed from a `WeightedAdd' layer; the output of this layer is passed from $3$ dense layers as in the combination of two modalities. Combining all three modalities has performed better than the remaining models in all the evaluation metrics. As observed during the experiments above, combining the information from the complementary modalities has led to better emotion recognition performance. Hence, the baselines and proposed model have been formulated with all three modalities and various information fusion mechanisms in Section \ref{sec:models}. %We started by using the three modalities, i.e., image, speech, and text, separately and alone to classify the emotion in the IIT-R MMEmoRec dataset. %The Image comes out to be the worst performer in terms of both, while the text is the best performer

%Actual Position: Elsewhere 
% \begin{table*}[!b]
% 		\centering	
% 		\caption{Ablation Studies.\vspace{-.05in}}
% 		\label{tab:res}
% 		\resizebox{.99\textwidth}{!}{%
% 			\begin{tabular}{l}
% 				\includegraphics[width=.8\textwidth]{results.jpg}
% 			\end{tabular}%
% 		} \vspace{.1in}
% 	\end{table*} 

\subsubsection{Ablation Study 2: Determining $k$ Values for KAAP}\label{sec:ablation3} 
\begin{figure}[!h]
	\centering
	\includegraphics[width=0.5\textwidth]{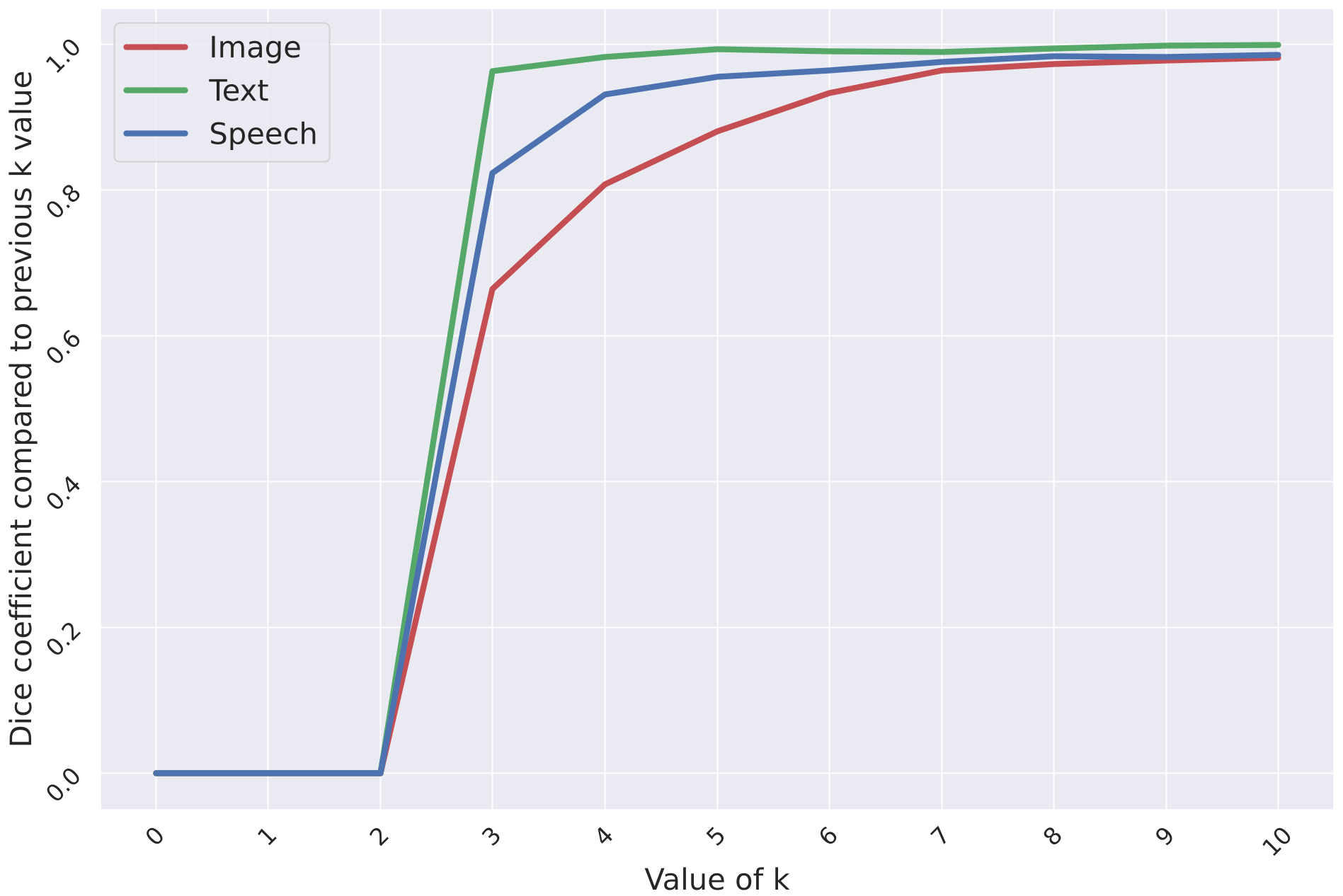}\vspace{-.1in}
	\caption{Ablation Study 2: Determining appropriate $k$ values.}\vspace{-.15in} 
	\label{fig:k_img}
\end{figure}

An in-depth ablation study has been conducted here to decide the value of $k$ used in Section \ref{sec:kaap_value}. The dice coefficient \cite{deng2018learning} is used to determine the best $k$ values. It measures the similarity of two data samples; the value of $1$ denotes that the two compared data samples are completely similar, whereas a value of $0$ denotes their complete dis-similarity. For each modality, KAAP values are calculated at $k \in \{2, 3 \dotsc, 10\}$. The dice coefficient is calculated for two adjacent $k$ values. For example, at $k=3$, the KAAP values at $k=2$ and $k=3$ are used to calculate the dice coefficient. This procedure was applied to all three modalities, and the results showing the effect of increasing $k$ are visualized in Fig.~\ref{fig:k_img}. For image \& speech, the value converges to $1$ at $k$ = $7$, while for text, the optimal value of k is $5$.  %It is an expected result as the image size is ($128$, $128$) much larger than any text containing $10$ to $20$ words.

% \begin{figure}[!h]
% 	\centering
% 	\includegraphics[width=0.49\textwidth]{fig_k_choice_img} 
% 	\caption{Determination of `k' value for Image modality.\vspace{-.3in}} 
% 	\label{fig:k_img}
% \end{figure}      

% \begin{figure}[!h]
% 	\centering
% 	\includegraphics[width=0.49\textwidth]{fig_k_choice_audio} 
% 	\caption{Determination of `k' value for Speech modality.\vspace{-.3in}} 
% 	\label{fig:k_speech}
% \end{figure}  

% \begin{figure}[!h]
% 	\centering
% 	\includegraphics[width=0.49\textwidth]{fig_k_choice_text} 
% 	\caption{Determination of `k' value for Text modality.\vspace{-.3in}} 
% 	\label{fig:k_text}
% \end{figure}  

\subsubsection{Ablation Study 3: Performance for Missing Modalities}\label{sec:ablation4} 
In real-life scenarios, some data samples may be missing information about one of the modalities. The VISTANet has been evaluated for such scenarios. We formulated four use cases with image, speech, text, or no modality missing and divided the test dataset into randomly selected equal parts accordingly. Then the information of the missing modality has been overridden to null, and VISTANet has been evaluated for emotion recognition. 

\begin{table}[!h]
\centering	
\caption{Ablation Study 3 on missing modalities. `Acc,' `P,' `R,' `F1,' \& `CK' denote accuracy, precision, recall, F1 \& CohenKappa scores. The highest values are marked in bold.\vspace{-.05in}}
\label{tab:miss_mod}
\resizebox{.37\textwidth}{!}{%
    \begin{tabular}{@{}lcccccc@{}}
    \toprule
    \textbf{Model}  & \textbf{Acc} & \textbf{F1} & \textbf{CK} & \textbf{P} & \textbf{R} \\ \midrule
    Missing Image   & 82.59  & 0.82 & 0.77 & 0.86 & 0.83\\
    Missing Speech  & 57.62  & 0.45 & 0.75 & 0.75 & 0.58 \\
    Missing Text    & 62.82  & 0.68 & 0.70 & 0.87 & 0.63
    \\\hdashline
    Missing None & \textbf{95.90} & \textbf{0.96} & \textbf{0.92} & \textbf{0.96} & \textbf{0.96}\\ \bottomrule
    \end{tabular}
}\vspace{-.05in}
\end{table}

As per Table \ref{tab:miss_mod}, the emotion recognition performance for missing no modality aligns with the results observed in Section\ref{sec:qualres}. Further, missing image modality information has caused the least dip in the performance. Moreover, the information from speech and text modalities combined has resulted in an accuracy of 82.59\%, whereas including all the modalities resulted in 95.90\% accuracy. These observations are consistent with Section \ref{sec:ablation2}, which reports that IER performance was lower than that of TER and SER.

\subsubsection{Ablation Study 4: Classwise Modality Pair Contribution}\label{sec:ablation5}
{
Fig.~\ref{fig:sample_res} provides sample‐level qualitative results. This study analyzes dataset‐wide, quantitative impact of each modality pair’s impact on each emotion. The six weights from the `WeightedAdd' layer that combines outputs $O_1$-$O_6$ in Eq. \ref{eq:eq2} are recorded for each test sample. Then the average is computed both overall and for each emotion class. Table~\ref{tab:pair_weights} shows the average weight contribution per pair, with weights L1-normalised to sum to 1.0 per row.\vspace{-.05in} %; higher values indicate greater influence in the final decision
}

\begin{table}[!h]
\centering
\caption{Ablation Study 4 on classwise contributions of modality pairs. Here, $P_i$, $P_s$, $P_t$ denote pre-trained and $S_i$, $S_s$, $S_t$ denote simpler subnetworks for image, speech, and text modalities. \vspace{-.05in}}
\label{tab:pair_weights}
\resizebox{.35\textwidth}{!}{%
\begin{tabular}{@{}lccccc@{}}
\toprule
\textbf{Pair} & \textbf{Overall} & \textbf{Angry} & \textbf{Happy} & \textbf{Hate} & \textbf{Sad}\\ \midrule
$\mathbf{P_i+S_s}$ & 0.13 & 0.14 & 0.11 & 0.13 & 0.12\\
$\mathbf{P_i+S_t}$ & 0.21 & 0.07 & 0.31 & 0.18 & 0.29\\
$\mathbf{P_s+S_i}$ & 0.11 & 0.19 & 0.07 & 0.07 & 0.09\\
$\mathbf{P_s+S_t}$ & 0.14 & 0.27 & 0.09 & 0.09 & 0.12\\
$\mathbf{P_t+S_i}$ & 0.26 & 0.19 & 0.30 & 0.31 & 0.27\\
$\mathbf{P_t+S_s}$ & 0.15 & 0.14 & 0.12 & 0.22 & 0.11\\ \bottomrule
\end{tabular}}
\vspace{-0.05in}
\end{table}

{
    These scores quantitatively indicate each modality pair’s contribution in enabling VISTANet to predict specific emotion classes. Speech \& text is the most contributing pair for `anger' class while text \& image influences `hate' the most. The `happy' and `sad' classes get influenced the most by image \& text pairs. %Angry (1% 93% 5%); Happy (49% 0% 49%); Hate (31% 1% 68%);	Sad (43% 13% 43%). 
}

\subsection{Discussion}
While many multimodal tasks prioritize a single input, VISTANet learns optimal fusion weights across all six modality pairs for balanced integration. {The bi-modal configuration uses two active modalities; in contrast, the tri-modal case with a missing modality still processes three inputs (two real and one zeroed), leading to different fusion dynamics and performance outcomes.} Leveraging the IIT R MMEmoRec dataset’s complementary cues, the network learns context-linked emotions and continuously corrects bias introduced by averaged unimodal labels. {Image is the most complex and diverse modality (wide variations in size, hue, saturation, angle, brightness, objects and backgrounds), whereas speech spectrograms and text embeddings are more structured and less diverse. Removing text or speech eliminates structural information the model expects, causing larger accuracy drops than zeroing the image (the image-only setup yields only 60.44\% accuracy, and the “Missing Image” case sees a smaller overall drop than omitting speech or text).} Human evaluators verified label consistency and audio quality. Ablation studies and missing-modality experiments confirm the benefit of integrating all signals. Finally, the KAAP technique quantifies each modality’s and feature’s influence and offers a tool for future advances in multimedia emotion analysis.

\section{Conclusions and future work}\label{sec:conclusion}	
The proposed system, VISTANet, performs emotion recognition by considering the information from the image, speech \& text modalities. It combines the information from these modalities in a hybrid manner of intermediate and late fusion and determines their weights automatically. It has resulted in better performance on including image, speech \& text modalities than including only one or two of these modalities. The proposed interpretability technique, KAAP, identifies each modality's contribution and important features toward predicting a particular emotion class. The future research plan includes transforming emotional content from one modality to another. We will also work on controllable emotion generation, where the output contains the desired emotional tone.

%%%%%%%%%%%%%%%%%%%%%%%%%%%%%%%%%%%%%%%%%%%%%%%%%%%%%%%%
%\section*{Declaration of Competing Interest}
% The authors declare that they have no known competing financial interests or personal relationships that could have appeared to influence the work reported in this paper.

%%%%%%%%%%%%%%%%%%%%%%%%%%%%%%%%%%%%%%%%%%%%%%%%%%%%%%%%
\section*{Acknowledgements}\label{sec:ack}
This work was initiated at the Machine Intelligence Lab, Indian Institute of Technology Roorkee, India, and extended at the Center for Machine Vision \& Signal Analysis, University of Oulu, Finland. It was supported by Research Council of Finland Profi 5 HiDyn fund (grant 24630111132) and Eudaimonia Institute of the University of Oulu. The authors also acknowledge the CSC-IT Center for Science, Finland, for providing computational resources. 

\ifCLASSOPTIONcaptionsoff
  \newpage
\fi

% =*=*=*=*=*=*=*=*=*=*=*=*=*=*=*=*=*=*=*=*=*=*=*=*=*=*=*=*=*=*
% Ref section
\bibliographystyle{IEEEtran} 
\bibliography{ref} 
\vspace{-.3in}

% =*=*=*=*=*=*=*=*=*=*=*=*=*=*=*=*=*=*=*=*=*=*=*=*=*=*=*=*=*=*
% biography section % For photo, change to {IEEEbiography}
\begin{IEEEbiography}
[{\includegraphics[width=1in,height=1.25in,clip,keepaspectratio]{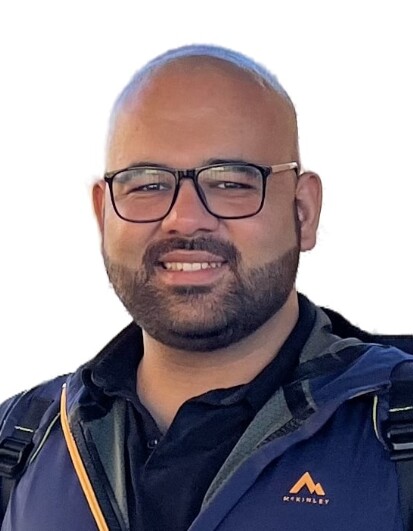}}]{Puneet~Kumar} (Member, IEEE) received his B.E. and M.E. degrees in Computer Science in 2014 and 2018, respectively, and his Ph.D. from the IIT Roorkee, India in 2022. He is a Postdoc at the University of Oulu, Finland and an Affiliated Scientist at the Meditation Research Program, USA. He works on Affective Computing, Multimodal \& Interpretable AI, Cognitive Neuroscience and Mental Health; has published in top journals \& conferences and received institute medal in M.E., best Ph.D. thesis award and several best paper awards. For more information, visit his webpage \url{www.puneetkumar.com}. \vspace{-.5in}
\end{IEEEbiography} 

\begin{IEEEbiography}[{\includegraphics[width=1in,height=1.25in,clip,keepaspectratio]{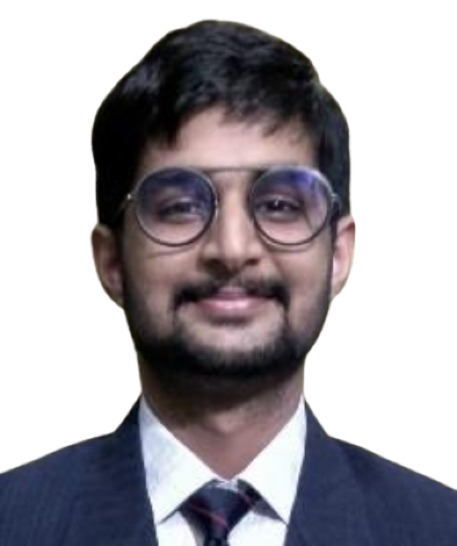}}]{Sarthak~Malik} received his B.Tech degree in Electrical Engineering from the Indian Institute of Technology Roorkee, India. He is currently a Data Scientist at MasterCard. He is an avid programmer and an active participant in coding competitions and development projects. His research focuses on Affective Computing, Computer Vision, and Interpretable AI. He has published in reputed journals and conferences and achieved notable recognitions, including a rank in the top 10 in the Geoffrey Hinton Fellowship hackathon. For more details, visit his webpage 
\href{https://www.linkedin.com/in/sarthak-malik-03777a190}{www.linkedin.com/in/sarthak-malik}.\vspace{-.35in}
\end{IEEEbiography}

\begin{IEEEbiography}[{\includegraphics[width=1in,height=1.25in,clip,keepaspectratio]{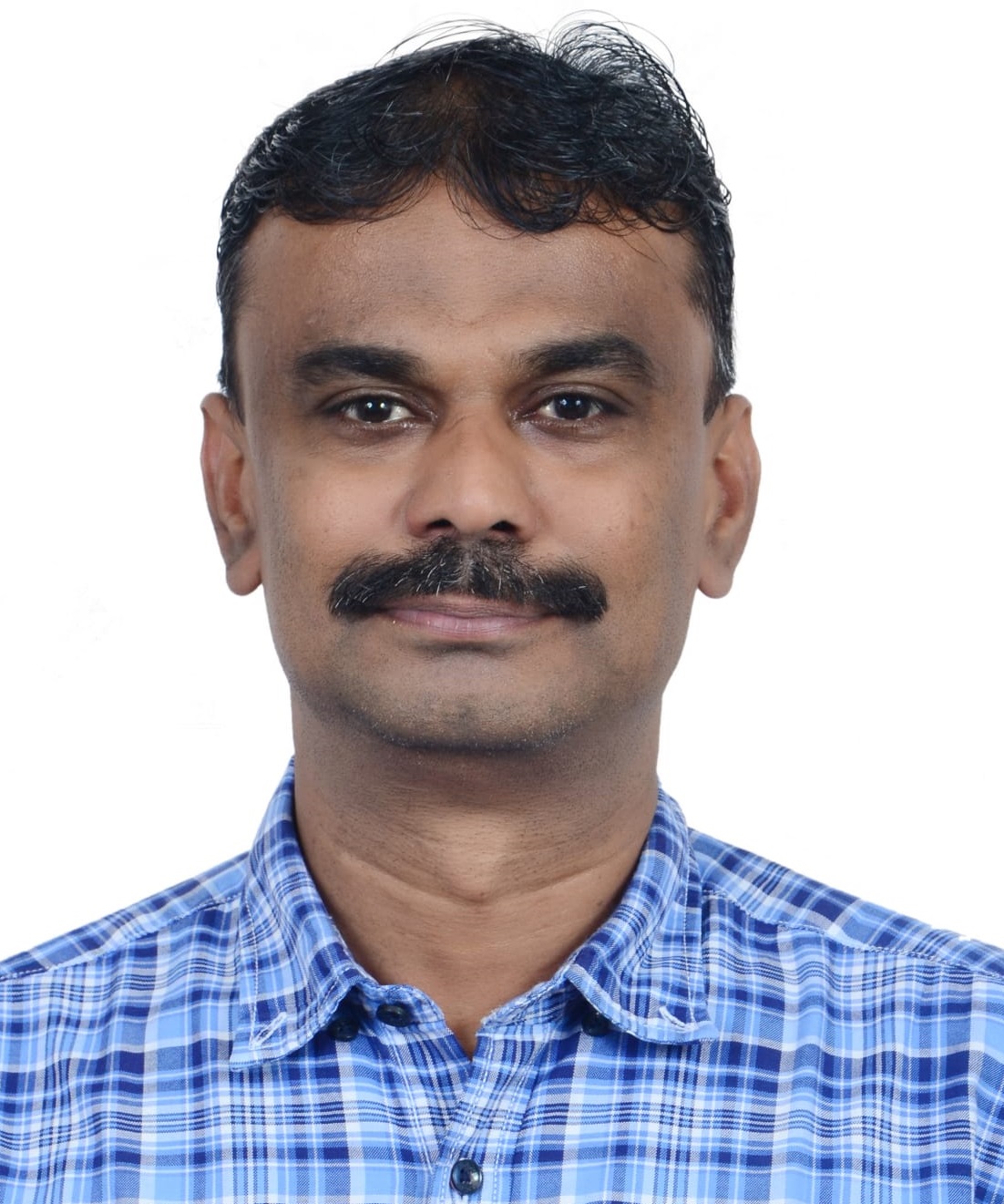}}]{Balasubramanian~Raman}
(Senior Member, IEEE) received his B.Sc. and M.Sc. degrees from the University of Madras in 1994 and 1996, respectively, and his Ph.D. from the IIT Madras in 2001. He is the Head and Chair Professor in the Department of Computer Science and a Joint Faculty in the Mehta Family School of Data Science \& AI at IIT Roorkee, India. He has published over 200 research papers in reputed journals and conferences. His research interests include Machine Learning, Image and Video Processing, Medical Imaging, Computer Vision, and Pattern Recognition. For more information, visit his webpage at \url{http://faculty.iitr.ac.in/cs/bala}. \vspace{-.35in}
\end{IEEEbiography}

\begin{IEEEbiography}[{\includegraphics[width=1in,height=1.25in,clip,keepaspectratio]{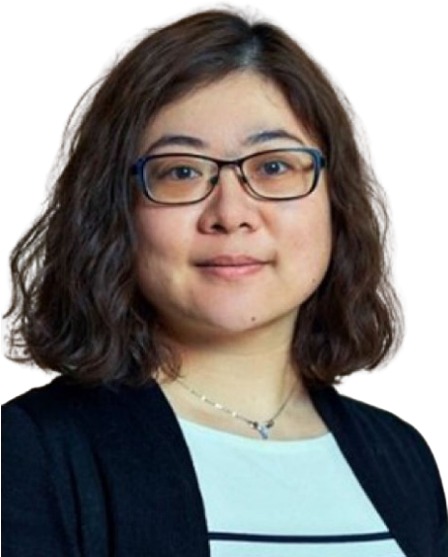}}]{Xiaobai~Li} (Senior Member, IEEE) received her B.Sc. and M.Sc. in 2004 and 2007, and her Ph.D. from the University of Oulu in 2017. She is a ZJU100 Professor at the State Key Lab of Blockchain and Data Security, Zhejiang University, and an Adjunct Professor at the University of Oulu. Her research covers Affective Computing, Facial and Micro-Expression Recognition, and Remote Physiological Signal Measurement. She has co-chaired international workshops at CVPR, ICCV, FG, and ACM MM, and is an Associate Editor for IEEE TMM, TCSVT, Frontiers in Psychology, and Image and Vision Computing. For details, visit her webpage \url{xiaobaili-uhai.github.io/}.
\end{IEEEbiography}

\end{document}